%% file: main.tex
\documentclass[10pt,twocolumn,letterpaper]{article}

\usepackage{cvpr}              

\usepackage{graphicx}
\usepackage{amssymb}
\usepackage{symmath}
\usepackage{booktabs}
\usepackage{multirow}
\usepackage{makecell}
\usepackage{amssymb}
\usepackage{pifont}

\usepackage{color, colortbl} 

\usepackage{tabu}           
\usepackage{adjustbox}
\usepackage{bm}

\input{math_commands}

\usepackage[pagebackref,breaklinks,colorlinks]{hyperref}
\usepackage[dvipsnames]{xcolor}
\newcommand{\layer}[1]{\ensuremath{\mathsf{#1}\xspace}}
\newcommand{\subsec}[1]{\noindent\textbf{#1}~~}

\usepackage[capitalize]{cleveref}
\crefname{section}{Sec.}{Secs.}
\Crefname{section}{Section}{Sections}
\Crefname{table}{Table}{Tables}
\crefname{table}{Tab.}{Tabs.}

\def\cvprPaperID{9404} 
\def\confName{CVPR}
\def\confYear{2022}

\newcommand{\papertitle}{DeepFace-EMD: Re-ranking Using Patch-wise Earth Mover's Distance \\Improves Out-Of-Distribution Face Identification}

\begin{document}


\title{\papertitle}
\author{Hai Phan$^{1,2}$\\
{\tt\small pthai1204@gmail.com}
\and
Anh Nguyen$^1$\\
{\tt\small anh.ng8@gmail.com}
\and
\hspace{1cm}$^1${Auburn University}\hspace{1cm}$^2${Carnegie Mellon University}
}

\twocolumn[{%
\maketitle
\begin{center}
    \centering
    {
    \small
    \begin{flushleft}
    \hspace{1.6cm}
    (a) LFW                 \hspace{2.3cm} 
    (b) Masked (LFW)        \hspace{1.7cm}
    (c) Sunglasses (LFW)    \hspace{1.5cm}
    (d) Profile (CFP)       \hfill
    \end{flushleft}
    }
    \includegraphics[scale=0.5]{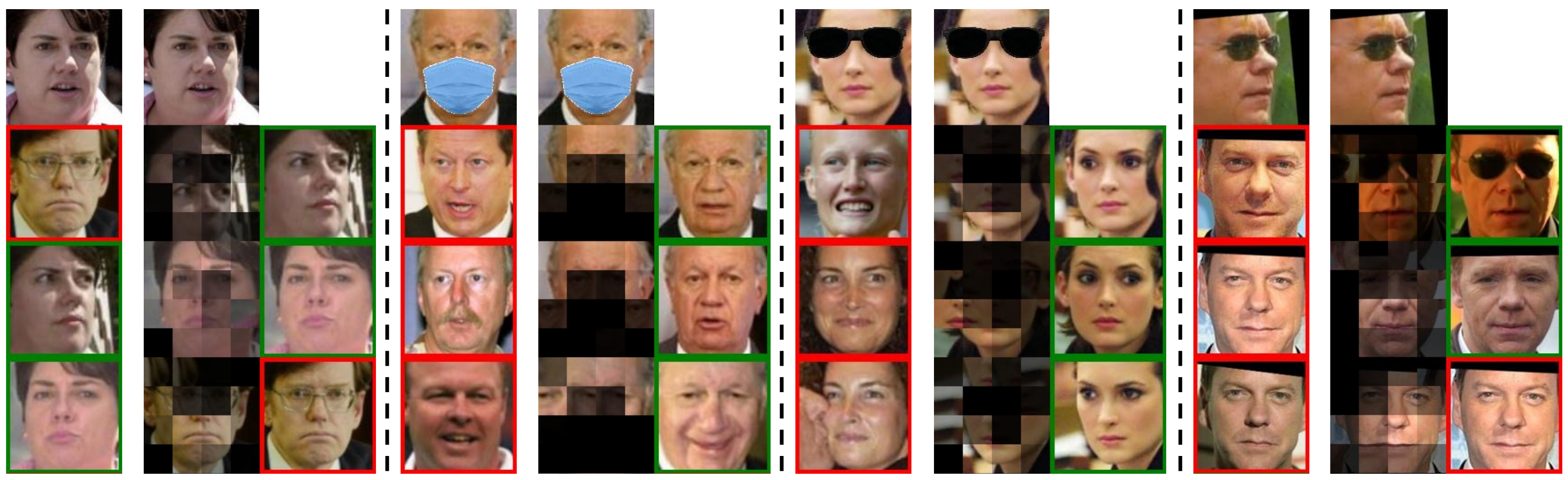}
    \vspace{-0.3cm}
    {
    \scriptsize
    \begin{flushleft}
    \hspace{0.7cm}
    Stage 1 \hspace{0.7cm} Flow \hspace{0.5cm} Stage 2
    \hspace{0.6cm}
    Stage 1 \hspace{0.7cm} Flow \hspace{0.5cm} Stage 2
    \hspace{0.6cm}
    Stage 1 \hspace{0.7cm} Flow \hspace{0.5cm} Stage 2
    \hspace{0.6cm}
    Stage 1 \hspace{0.7cm} Flow \hspace{0.5cm} Stage 2
    \hfill
    \end{flushleft}
    \vspace{-0.3cm}
    }
    \captionof{figure}{
    Traditional face identification ranks gallery images based on their cosine distance with the query (top row) at the image-level embedding, which yields large errors upon out-of-distribution changes in the input (\eg masks or sunglasses; b--d).
    We find that re-ranking the top-$k$ shortlisted faces from Stage 1 (leftmost column) using their patch-wise EMD similarity w.r.t. the query substantially improves the precision (Stage 2) on challenging cases (b--d).
    The ``Flow'' visualization intuitively shows the patch-wise reconstruction of the query face using the most similar patches (\ie highest flow) from the retrieved face.
    See \cref{fig:face_vis_44} for a full figure with the top-5 candidates.
    }\label{fig:teaser}
\end{center}%
}]

\begin{abstract}

Face identification (FI) is ubiquitous and drives many high-stake decisions made by the law enforcement.
A common FI approach compares two images by taking the cosine similarity between their image embeddings.
Yet, such approach suffers from poor out-of-distribution (OOD) generalization to new types of images (e.g., when a query face is masked, cropped or rotated) not included in the training set or the gallery.
Here, we propose a re-ranking approach that compares two faces using the Earth Mover's Distance on the deep, spatial features of image patches.
Our extra comparison stage explicitly examines image similarity at a fine-grained level (e.g., eyes to eyes) and is more robust to OOD perturbations and occlusions than traditional FI.
Interestingly, without finetuning feature extractors, our method consistently improves the accuracy on all tested OOD queries: masked, cropped, rotated, and adversarial while obtaining similar results on in-distribution images.

\end{abstract}

\section{Introduction}
\label{sec:intro}

Who shoplifted from the Shinola luxury store in Detroit \cite{lawsuit2021facial}? 
Who are you to receive unemployment benefits \cite{MiaSato.2021} or board an airplane \cite{airplane}?
Face identification (FI) today is behind the answers to such life-critical questions. 
Yet, the technology can make errors, leading to severe consequences, \eg people wrongly denied of unemployment benefits \cite{MiaSato.2021} or falsely arrested \cite{lawsuit2021facial,newjersey2021arrested,detroit2021arrested,michigan2021arrested}. 
Identifying the person in a single photo remains challenging because, in many cases, the problem is a zero-shot and ill-posed image retrieval task. 
First, a deep feature extractor may not have seen a normal, non-celebrity person before during training. 
Second, there may be too few photos of a person in the database for FI systems to make reliable decisions. 
Third, it is harder to identify when a face in the wild (\eg from surveillance cameras) is occluded \cite{sun2015deeply,qiu2021end2end} (\eg wearing masks), distant or cropped, yielding a new type of photo not in both the training set of deep networks and the retrieval database---i.e., out-of-distribution (OOD) data.
For example, face verification accuracy may notoriously drop significantly (from 99.38\% to 81.12\% on LFW) given an occluded query face (\cref{fig:teaser}b--d) \cite{qiu2021end2end} or adversarial queries \cite{zhong2020towards,amos2016openface}.

In this paper, we propose to evaluate the performance of state-of-the-art facial feature extractors (ArcFace \cite{deng2018arcface}, CosFace \cite{wang2018cosface}, and FaceNet \cite{schroff2015facenet}) on OOD face \emph{identification} tests. 
That is, our main task is to recognize the person in a query image given a gallery of known faces.
Besides in-distribution (ID) query images, we also test FI models on OOD queries that contain (1) common occlusions, \ie random crops, faces with masks or sunglasses; and (2) adversarial perturbations \cite{zhong2020towards}.
Our main findings are:\footnote{Code, demo and data are available at \url{https://github.com/anguyen8/deepface-emd}}

\begin{itemize}

    \item Interestingly, the OOD accuracy can be substantially improved via a 2-stage approach (see \cref{fig:framework}): First, identify a set of the most globally-similar faces from the gallery using cosine distance and then, re-rank these shortlisted candidates by comparing them with the query at the patch-embedding level using the Earth Mover's Distance (EMD) \cite{rubner2000earth} (\cref{sec:ablation_study} \& \cref{sec:exper_result}). 
    
    \item Across three different models (ArcFace, CosFace, and FaceNet), our re-ranking approach consistently improves the original precision (under all metrics: P@1, R-Precision, and MAP@R) \emph{without finetuning} (\cref{sec:exper_result}). 
    That is, interestingly, the spatial features extracted from these models can be leveraged to compare images patch-wise (in addition to image-wise) and further improve FI accuracy.
    
    \item On masked images \cite{wang2021mlfw}, our re-ranking method (no training) rivals the ArcFace models finetuned directly on masked images (\cref{sec:com_finetune}). 
    
\end{itemize}

To our knowledge, our work is the first to demonstrate the remarkable effectiveness of EMD for comparing OOD, occluded and adversarial images at the deep feature level.

\section{Methods}
\label{sec:methods}

\subsection{Problem formulation}
To demonstrate the generality of our method, we adopt the following simple FI formulation as in \cite{zhao2021towards,deng2018arcface,Liu_2017_CVPR}: Identify the person in a query image by ranking all gallery images based on their pair-wise similarity with the query.
After ranking (Stage 1) or re-ranking (Stage 2), we take the identity of the top-1 nearest image as the predicted identity.

\subsec{Evaluation} Following \cite{zhao2021towards,musgrave2020metric}, we use three common evaluation metrics: Precision@1 (P@1), R-Precision (RP), and MAP@R (M@R).
See their definitions in Sec.~B1 in \cite{zhao2021towards}.

\subsection{Networks}

\paragraph{Pre-trained models}
We use three state-of-the-art PyTorch models of ArcFace, FaceNet, and CosFace pre-trained on CASIA \cite{yi2014learning}, VGGFace2 \cite{cao2018vggface2}, and CASIA, respectively.
Their architectures are ResNet-18 \cite{he2016identity}, Inception-ResNet-v1 \cite{szegedy2017inception}, and 20-layer SphereFace \cite{Liu_2017_CVPR}, respectively.
See \cref{sec:supp_pretrained_models} for more details on network architectures and implementation in PyTorch.

\subsec{Image pre-processing} 
For all networks, we align and crop input images following the 3D facial alignment in \cite{Bhagavatula2017FasterTR} (which uses 5 reference points, 0.7 and 0.6 crop ratios for width and height, and Similarity transformation).
All images shown in this paper (\eg \cref{fig:teaser}) are pre-processed.
Using MTCNN, the default pre-processing of all three networks, does not change the results substantially (Sec.~\ref{sec:mtcnn}).

\subsection{2-stage hierarchical face identification}

\begin{figure*}
    \centering
    \includegraphics[width=0.9\textwidth]{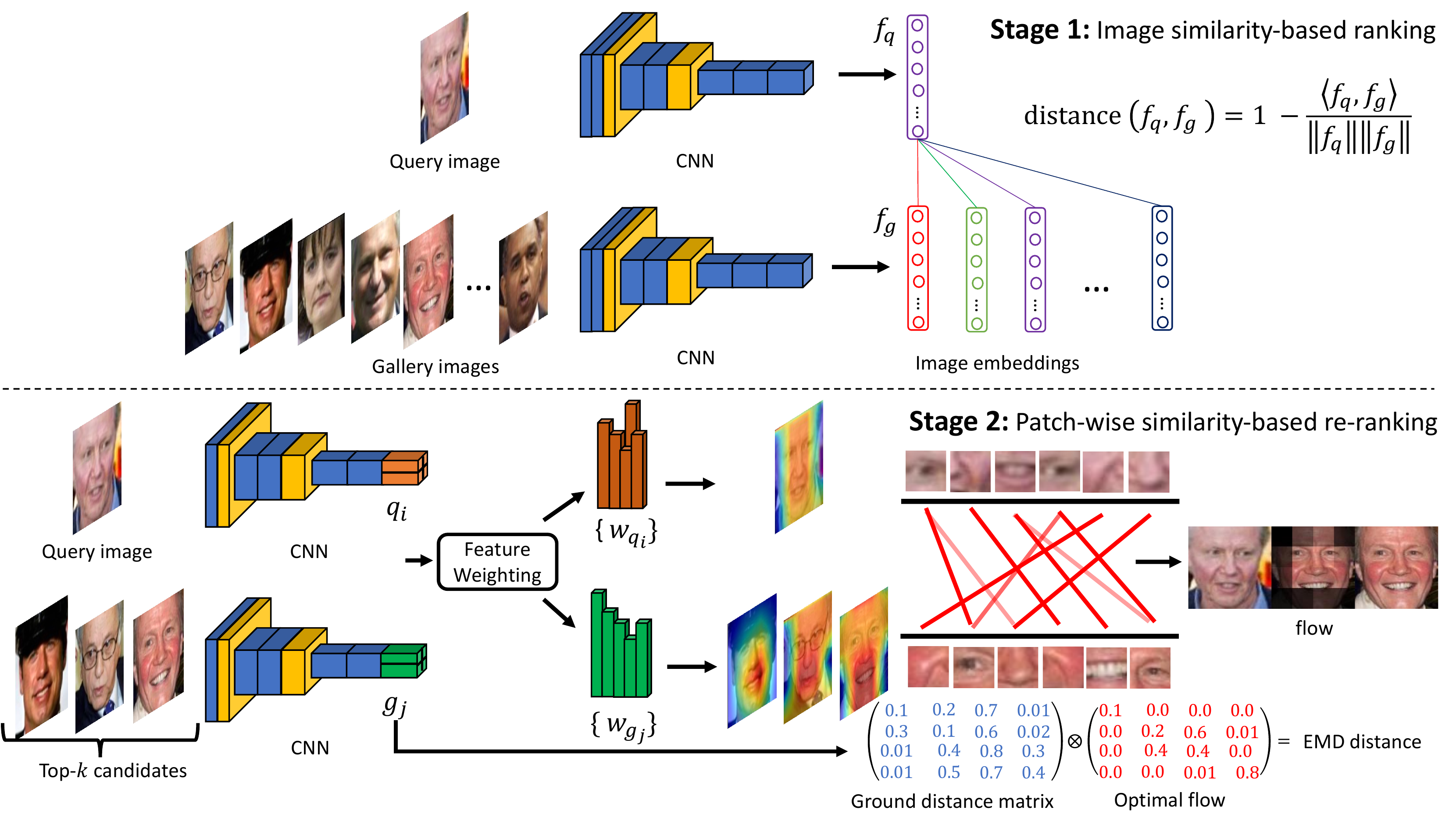}
    \caption{
    Our 2-stage face identification pipeline. 
    Stage 1 ranks gallery images based on their cosine distance with the query face at the image-embedding level.
    Stage 2 then re-ranks the top-$k$ shortlisted candidates from Stage 1 using EMD at the patch-embedding level.
    } 
    \vspace*{-0.5cm}
    \label{fig:framework}
\end{figure*}

\subsec{Stage-1: Ranking} 
A common 1-stage face identification
\cite{Liu_2017_CVPR,schroff2015facenet,wang2018cosface} ranks gallery images based on their pair-wise cosine similarity with a given query in the last-linear-layer feature space of a pre-trained feature extractor (\cref{fig:framework}).
Here, our image embeddings are extracted from the \emph{last linear} layer of all three models and are all $\in \sR^{512}$.

\subsec{Stage-2: Re-ranking}
We re-rank the top-$k$ (where the optimal $k$ = 100) candidates from Stage 1 by computing the patch-wise similarity for an image pair using EMD.
Overall, we compare faces in \textbf{two hierarchical stages} (\cref{fig:framework}), first at a coarse, image level and then a fine-grained, patch level.

Via an ablation study (\cref{sec:ablation_study}), we find our 2-stage approach (a.k.a. DeepFace-EMD) more accurate than Stage 1 alone (\ie no patch-wise re-ranking) and also Stage 2 alone (\ie sorting the entire gallery using patch-wise similarity).

\subsection{Earth Mover's Distance (EMD)}
\label{sec:emd}

EMD is an edit distance between two set of weighted objects or distributions \cite{rubner2000earth}. 
Its effectiveness was first demonstrated in measuring pair-wise image similarity based on color histograms and texture frequencies \cite{rubner2000earth} for image retrieval. 
Yet, EMD is also an effective distance between two text documents \cite{kusner2015word}, probability distributions (where EMD is equivalent to Wasserstein, \ie Mallows distance) \cite{levina2001earth}, and distributions in many other domains \cite{peng2006using,lupu2017new,kumar2017earth}.
Here, we propose to harness EMD as a distance between two faces, \ie two sets of weighted facial features.

Let $\gQ = \{ (q_1, w_{q_1}), ..., (q_N, w_{q_N}) \}$ be a set of $N$ (facial feature, weight) pairs describing a query face where $q_i$ is a feature (\eg left eye or nose) and the corresponding $w_{q_i}$ indicates how important the feature $q_i$ is in FI.
The \emph{flow} between $\gQ$ and the set of weighted features of a gallery face $\gG = \{ (g_1, w_{g_1}), ..., (g_N, w_{g_N}) \}$ is any matrix $\mF = (f_{ij}) \in \sR^{N\times N}$.
Intuitively, $f_{ij}$ is the amount of importance weight at $q_i$ that is matched to the weight at $g_j$.
Let $d_{ij}$ be a ground distance between ($q_i$, $g_j$) and $\mD = (d_{ij}) \in \sR^{N\times N}$ be the ground distance matrix of all pair-wise distances. 

We want to find an optimal flow $\mF$ that minimizes the following cost function, \ie the sum of weighted pair-wise distances across the two sets of facial features:

\vspace*{-0.6cm}
\begin{align}
    \label{eq:emd_opt}
    \text{COST} (\gQ, \gG, \mF) = 
    \sum_{i=1}^N\sum_{j=1}^N d_{ij} f_{ij}
\end{align}
\vspace*{-0.5cm}
\begin{align}
\text{s.t.} \quad \quad \quad f_{ij} &\ge 0 \\
\sum_{j=1}^N f_{ij} &\le w_{q_i}, \text{and}
\sum_{i=1}^N f_{ij} \le w_{g_j}, i,j \in [1,N] \\
\sum_{j=1}^N \sum_{i=1}^N f_{ij} &= \min \left (\sum_{j=1}^N w_{g_j}, \sum_{i=1}^N w_{q_i} \right ).
\label{eq:total_flow}
\end{align}

\noindent As in \cite{zhao2021towards,zhang2020deepemdv2}, we normalize the weights of a face such that the total weights of features is 1 \ie $\sum^N_{i=1}w_{q_i} = \sum^N_{j=1}w_{g_j} = 1$, which is also the total flow in \cref{eq:total_flow}.
Note that EMD is a metric iff two distributions have an equal total weight and the ground distance function is a metric \cite{cohen1999finding}.

We use the iterative Sinkhorn algorithm \cite{NIPS2013_af21d0c9} to efficiently solve the linear programming problem in \cref{eq:emd_opt}, which yields the final EMD between two faces $\gQ$ and $\gG$.

\subsec{Facial features}
In image retrieval using EMD, a set of features $\{ q_i \}$ can be a collection of dominant colors \cite{rubner2000earth}, spatial frequencies \cite{rubner2000earth}, or a histogram-like descriptor based on the local patches of reference identities \cite{wang2012supervised}.
Inspired by \cite{wang2012supervised}, we also divide an image into a grid but we take the embeddings of the local patches from the last convolutional layers of each network.
That is, in FI, face images are aligned and cropped such that the entire face covers most of the image (see \cref{fig:teaser}a).
Therefore, without facial occlusion, every image patch is supposed to contain useful identity information, which is in contrast to natural photos \cite{zhang2020deepemdv2}.

Our grid sizes $H\times W$ for ArcFace, FaceNet, and CosFace are respectively, 8$\times$8, 3$\times$3, and 6$\times$7, which are the corresponding spatial dimensions of their last convolutional layers (see definitions of these layers in \cref{sec:supp_pretrained_models}).
That is, each feature $q_i$ is an embedding of size 1$\times$1$\times C$ where $C$ is the number of channels (\ie 512, 1792, and 512 for ArcFace, FaceNet, and CosFace, respectively).

\subsec{Ground distance}
Like \cite{zhang2020deepemdv2,zhao2021towards}, we use cosine distance as the ground distance $d_{ij}$ between the embeddings ($q_i$, $g_j$) of two patches:

\vspace*{-0.5cm}
\begin{equation}
    \begin{aligned}
    d_{ij} = 1 - \frac{\left \langle q_i, g_j \right \rangle}{\left \|q_i \right \| \left \| g_j \right \|}
    \end{aligned}
\end{equation}

where $\left \langle . \right \rangle$ is the dot product between two feature vectors.

\subsection{Feature weighting}
\label{sec:fea_selec}

EMD in our FI intuitively is an optimal plan to match all weighted features across two images.
Therefore, how to weight features is an important step. 
Here, we thoroughly explore five different feature-weighting techniques for FI.

\subsec{Uniform}
Zhang et al. \cite{zhang2020deepemdv2} found that it is beneficial to assign lower weight to less informative regions (\eg background or occlusion) and higher weight to discriminative areas (\eg those containing salient objects).
Yet, assigning an equal weight to all $N = H \times W$ patches is worth testing given that background noise is often cropped out of the pre-processed face image (\cref{fig:teaser}):

\begin{equation}
    \begin{aligned}
    w_{q_i} = w_{g_i} = \frac{1}{N}, \text{where} \ 1 \le k \le N
    \end{aligned}
\end{equation}

\begin{figure}[t]
    {
        \small
		\begin{flushleft}
			\hskip 0.25in (a) SC
			\hskip 0.45in (b) APC
			\hskip 0.75in (c) LMK
		\end{flushleft}
	}
	\vskip -0.1in
    \centering
    \includegraphics[width=0.5\textwidth]{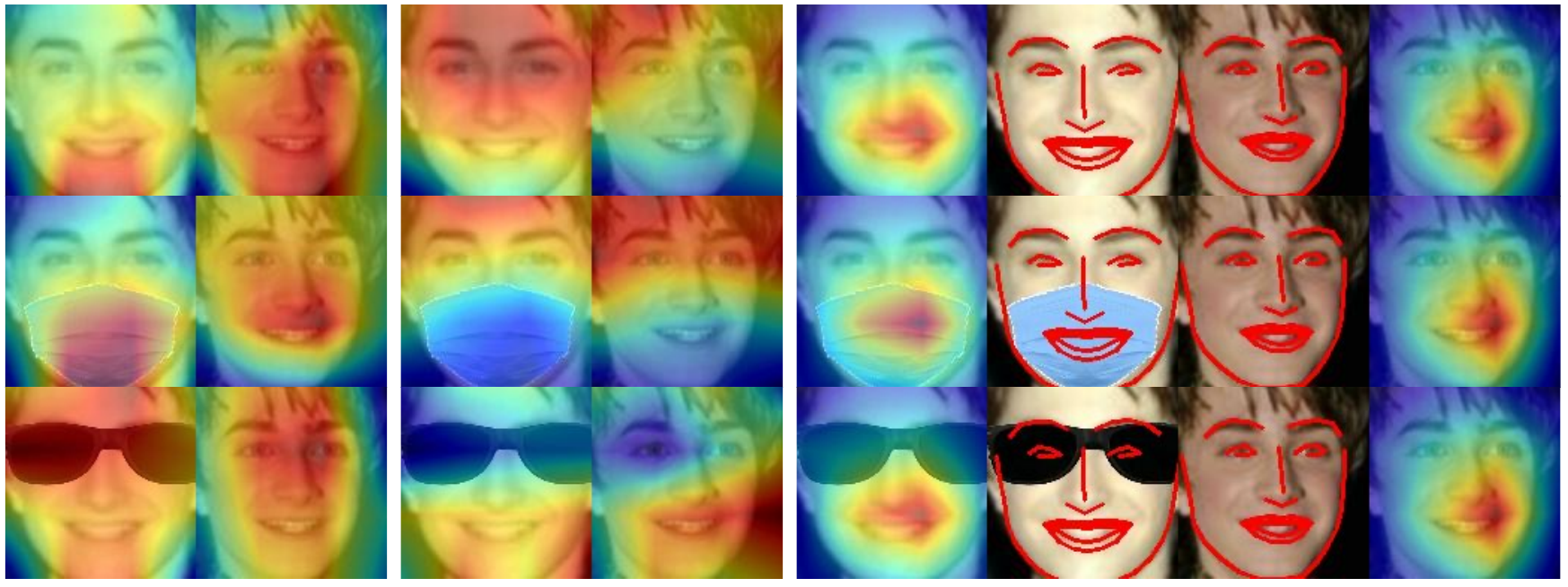}
    \caption{
    The results of assigning weights to 4$\times$4 patches for ArcFace under three different techniques.
    Based on per-patch density of detected landmarks (\textcolor{red}{- - -}), LMK (c) often assigns higher weight to the center of a face (regardless of occlusions).
    In contrast, SC and APC assign higher weight to patches of higher patch-wise and patch-vs-image similarity, respectively.
    APC tends to down-weight a facial feature (\eg blue regions around sunglasses or mouth) if its corresponding feature is occluded in the other image (b).
    In contrast, SC is insensitive to occlusions (a).
    See ~\cref{fig:heatmap} \& ~\cref{fig:heatmap_flows_single} for more heatmap examples of feature weighting.
    }
    \label{fig:feature_weighting}
\end{figure}

\subsec{Average Pooling Correlation (APC)} 
Instead of uniformly weighting all patch embeddings, an alternative from \cite{zhang2020deepemdv2} would be to weight a given feature $q_i$ proportional to its correlation to the entire other \emph{image} in consideration.
That is, the weight $w_{q_i}$ would be the dot product between the feature $q_i$ and the average pooling output of all embeddings $\{g_j\}_1^N$ of the gallery image:

\vspace*{-0.5cm}
\begin{align}
w_{q_i} = \max \big(0, \langle  q_i,  ~\frac{\sum_j^{N} g_j}{N} \rangle \big), 
w_{g_j} = \max \big (0, \langle g_j, \frac{\sum_i^{N} q_i}{N} \rangle \big)
\end{align}

where $\max(.)$ keeps the weights always non-negative.
APC tends to assign near-zero weight to occluded regions and, interestingly, also minimizes the weight of eyes and mouth in a non-occluded gallery image (see \cref{fig:feature_weighting}b; blue shades around both the mask and the non-occluded mouth).

\subsec{Cross Correlation (CC)}
APC \cite{zhang2020deepemdv2} is different from CC introduced in \cite{zhao2021towards}, which is the same as APC except that CC uses the output vector from the last linear layer (see \href{https://github.com/wl-zhao/DIML/blob/e0832b591f4985685c0ecfde89e836eac356f3f6/evaluation/eval_diml.py#L37}{code}) instead of the global average pooling vector in APC.

\subsec{Spatial Correlation (SC)} 
While both APC  and CC ``summarize'' an entire other gallery image into a vector first, and then compute its correlation with a given patch $q_i$ in the query.
In contrast, an alternative, inspired by \cite{stylianouSimVis2019}, is to take the sum of the cosine similarity between the query patch $q_i$ and every patch in each gallery image $\{ g_j \}_1^N$:


\vspace*{-0.5cm}
\begin{align}
\label{eq:SC}
    w_{q_i} = \max \big(0, \sum_j^{N}\frac{ \left \langle q_i,  g_j \right \rangle }{\| q_i \| \| g_j\|} \big), 
    w_{g_j} = \max \big(0, \sum_i^{N}\frac{ \left \langle q_i,  g_j \right \rangle }{\| q_i \| \| g_j\|} \big)
\end{align}

We observe that SC often assigns a higher weight to occluded regions \eg,  masks and sunglasses (\cref{fig:feature_weighting}b).

\subsec{Landmarking (LMK)} 
While the previous three techniques adaptively rely on the image-patch similarity (APC, CC) or patch-wise similarity (SC) to weight a given patch embedding, their considered important points may or may not align with facial landmarks, which are known to be important for many face-related tasks.
Here, as a baseline for APC, CC, and SC, we use \layer{dlib} \cite{dlib09} to predict 68 keypoints in each face image (see \cref{fig:feature_weighting}c) and weight each patch-embedding by the density of the keypoints inside the patch area.
Our LMK weight distribution appears Gaussian-like with the peak often right below the nose (\cref{fig:feature_weighting}c).

\section{Ablation Studies}
\label{sec:ablation_study}

We perform three ablation studies to rigorously evaluate the key design choices in our 2-stage FI approach: 
(1) Which feature-weighting techniques to use (\cref{sec:eff_fea})?
(2) re-ranking using both EMD and cosine distance (\cref{sec:eff_alpha}); and
(3) comparing patches or images in Stage 1 (\cref{sec:emd_stage_1}).

\subsec{Experiment}
For all three experiments, we use ArcFace to perform FI on both LFW \cite{yi2014learning} and LFW-crop.
For \textbf{LFW}, we take all 1,680 people who have $\geq$2 images for a total of 9,164 images.
When taking each image as a query, we search in a gallery of the remaining 9,163 images.
For the experiments with \textbf{LFW-crop}, we use all 13,233 original LFW images as the gallery. 
To create a query set of 13,233 cropped images, we clone the gallery and crop each image randomly to its 70\% and upsample it back to the original size of 128$\times$128 (see examples in \cref{fig:face_vis}d).
That is, LFW-crop tests identifying a cropped (\ie close-up, and misaligned) image given the unchanged LFW gallery.
LFW and LFW-crop tests offer contrast insights (ID vs. OOD).

In Stage 2, \ie re-ranking the top-$k$ candidates, we test different values of $k \in \{100, 200, 300\}$ and do not find the performance to change substantially.
At $k = 100$, our 2-stage precision is already close to the maximum precision of 99.88 under a perfect re-ranking (see \cref{tab:eff_fea}a; Max prec.).

\subsection{Comparing feature weighting techniques}
\label{sec:eff_fea}

Here, we evaluate the precision of our 2-stage FI as we sweep across five different feature-weighting techniques and two grid sizes (8$\times$8 and 4$\times$4).
In an 8$\times$8 grid, we observe that some facial features such as the eyes are often split in half across two patches (see \cref{fig:face_vis_88}), which may impair the patch-wise similarity.
Therefore, for each weighting technique, we also test average-pooling the 8$\times$8 grid into 4$\times$4 and performing EMD on the resultant 16 patches.

\subsec{Results}
First, we find that, on LFW, our image-similarity-based techniques (APC, SC) outperform the LMK baseline (\cref{tab:eff_fea}a) despite not using landmarks in the weighting process, verifying the effectiveness of adaptive, similarity-based weighting schemes.

Second, interestingly, in FI, we find that Uniform, APC, and SC all outperform the CC weighting proposed in \cite{zhao2021towards,zhang2020deepemdv2}.
This is in stark contrast to the finding in \cite{zhao2021towards} that CC is better than Uniform (perhaps because face images do not have background noise and are close-up).
Furthermore, using the global average-pooling vector from the channel (APC) substantially yields more useful spatial similarity than the last-linear-layer output as in CC implementation (\cref{tab:eff_fea}b; 96.16 vs. 91.31 P@1).

Third, surprisingly, despite that a patch in a 8$\times$8 grid does not enclose an entire, fully-visible facial feature (e.g. an eye), all feature-weighting methods are on-par or better on an 8$\times$8 grid than on a 4$\times$4 (\eg \cref{tab:eff_fea}b; APC: 96.16 vs. 95.32).
Note that the optimal flow visualized in a 4$\times$4 grid is more interpretable to humans than that on a 8$\times$8 grid (compare \cref{fig:teaser} vs. \cref{fig:face_vis_88}).

Fourth, across all variants of feature weighting, our 2-stage approach consistently and \emph{substantially} outperforms the traditional Stage 1 alone on LFW-crop, suggesting its robust effectiveness in handling OOD queries.

Fifth, under a perfect re-ranking of the top-$k$ candidates (where $k=100$), there is only 1.4\% headroom for improvement upon Stage 1 alone in LFW (\cref{tab:eff_fea}a; 98.48 vs. 99.88) while there is a large $\sim$12\% headroom in LFW-crop (\cref{tab:eff_fea}a; 87.35 vs. 98.71).
Interestingly, our re-ranking results approach the upperbound re-ranking precision (\eg \cref{tab:eff_fea}b; 96.26 of Uniform vs. 98.71 Max prec. at $k=100$).

\begin{table}[t]
\small
\centering
\resizebox{7cm}{!}{%
\begin{tabular}{ c|r|c|c|c } 
\hline
ArcFace & Method & P@1 & RP & M@R \\
\hline

\multirow{8}{*}{\thead{LFW \\ vs. \\ LFW \\ (a)}} & Stage 1 alone ~\cite{deng2018arcface} & 98.48 & 78.69 & 78.29\\ \cline{2-5}
&\cellcolor{yellow} Max prec. at $k = 100$ & \cellcolor{yellow} 99.88 & \cellcolor{yellow} 81.32 &  \cellcolor{yellow} - \\ \cline{2-5}
&CC \cite{zhao2021towards} ($8\times 8$) & 98.42 & 78.35 & 77.91  \\
&CC \cite{zhao2021towards} ($4\times 4$)  & 81.69 & 76.29 & 72.47  \\ \cline{2-5}
&APC ($8\times 8$) & \textbf{98.60 }& 78.63 & 78.23  \\
&APC ($4\times 4$) & \textbf{98.54} & 78.57 & 78.16 \\
&Uniform ($8 \times 8$) & \textbf{98.66} & \textbf{78.73} & \textbf{78.35} \\
&Uniform ($4 \times 4$) & \textbf{98.63} & \textbf{78.72} & \textbf{78.33} \\
&SC ($8 \times 8$) & \textbf{98.66} & \textbf{78.74} & \textbf{78.35}\\
&SC ($4 \times 4$) & \textbf{98.65} & \textbf{78.72} & \textbf{78.33}\\ \cline{2-5}
&LMK ($8 \times 8$) & 98.35 & 78.43 & 77.99 \\
&LMK ($4 \times 4$) & 98.31 & 78.38 & 77.90 \\

\hline
\multirow{8}{*}{\thead{LFW-crop \\ vs. \\ LFW \\ (b)}} & Stage 1 alone ~\cite{deng2018arcface} & 87.35 & 71.38 & 69.04\\ \cline{2-5}
& \cellcolor{yellow} Max prec. at $k = 100$ & \cellcolor{yellow} 98.71 & \cellcolor{yellow} 89.13 & \cellcolor{yellow} - \\ \cline{2-5}
&CC~\cite{zhao2021towards} ($8\times 8$) & 91.31 & 72.33 & 70.00  \\
&CC~\cite{zhao2021towards} ($4\times 4$) & 63.12 & 56.03 & 51.00  \\ \cline{2-5}
&APC ($8\times 8$) & \textbf{96.16}& \textbf{76.60} & \textbf{74.57}  \\
&APC ($4\times 4$) & \textbf{95.32} & \textbf{75.37} & \textbf{73.25} \\
&Uniform ($8\times 8$) & \textbf{96.26} & \textbf{78.08} & \textbf{76.25} \\
&Uniform ($4\times 4$) & \textbf{95.53} & \textbf{77.15} & \textbf{75.29} \\
&SC ($8\times 8$) & \textbf{96.19} & \textbf{78.05} & \textbf{76.20}\\
&SC ($4\times 4$) & \textbf{95.42} & \textbf{77.12} & \textbf{75.25}\\
\hline
\end{tabular}
}
\caption{Comparison of five feature-weighting techniques for ArcFace \cite{deng2018arcface} patch embeddings on LFW ~\cite{yi2014learning} and LFW-crop datasets. 
Performance is often slightly better on a 8$\times$8 grid than on a 4$\times$4.
Our 2-stage approach consistently outperforms the vanilla Stage 1 alone and approaches closely the maximum re-ranking precision at $k=100$.
}
\label{tab:eff_fea}
\end{table}

\subsection{Re-ranking using both EMD \& cosine distance}
\label{sec:eff_alpha}

We observe that for some images, re-ranking using patch-wise similarity at Stage 2 does not help but instead hurt the accuracy.
Here, we test whether linearly combining EMD (at the \emph{patch}-level embeddings as in Stage 2) and cosine distance (at the \emph{image}-level embeddings as in Stage 1) may improve \emph{re-ranking} accuracy further (vs. EMD alone).

\subsec{Experiment}
We use the grid size of 8$\times$8, \ie the better setting from the previous ablation study (\cref{sec:eff_fea}).
For each pair of images, we linearly combine their patch-level EMD ($\theta_\textrm{EMD}$) and the image-level cosine distance ($\theta_{\textrm{Cosine}}$) as:

\vspace{-0.5cm}
\begin{align}
    \theta = \alpha \times \theta_\textrm{EMD} + (1 - \alpha) \times \theta_{\textrm{Cosine}}
\label{eq:alpha}
\end{align}



\begin{figure}[h!] 
    \centering 
    \includegraphics[width=1.0\linewidth]{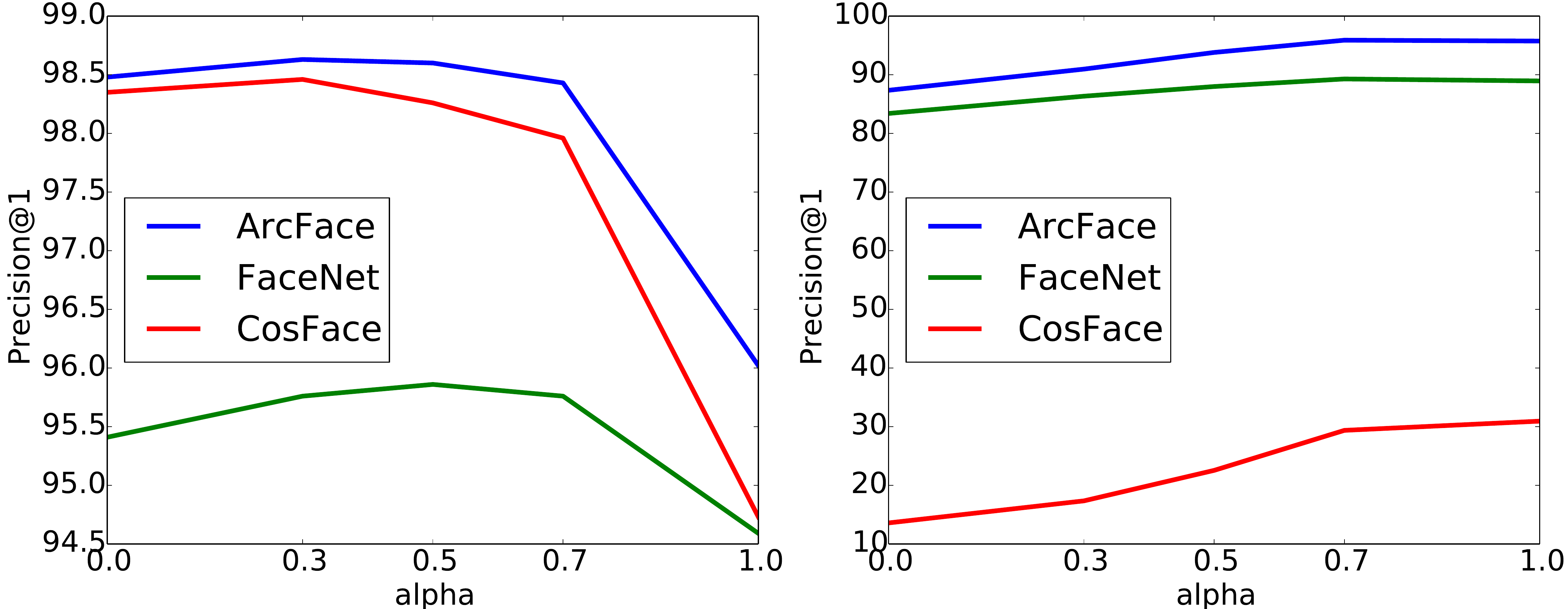}
    {
        \vspace*{-0.6cm}
        \small
		\begin{flushleft}
			\hskip 1.7cm (a) LFW
			\hskip 2.5cm (b) LFW-crop
		\end{flushleft}
	}
	\vskip -0.2in
\caption{
The P@1 of our 2-stage FI when sweeping across $\alpha$ for linearly combining EMD and cosine distance on LFW (a) and LFW-crop images (b) when using APC feature weighting. 
Trends are similar for all other feature-weighting methods (see \cref{fig:supp_alpha_effect}).
}
\label{fig:alpha_effect}
\end{figure}

Sweeping across  $\alpha \in \{0, 0.3, 0.5, 0.7, 1\}$, we find that changing $\alpha$ has a marginal effect on the P@1 on LFW. 
That is, the P@1 changes in [95, 98.5] with the lowest accuracy being 95 when EMD is exclusively used, \ie $\alpha = 1$ (see \cref{fig:alpha_effect}a).
In contrast, for LFW-crop, we find the accuracy to monotonically increases as we increase $\alpha$ (\cref{fig:alpha_effect}b).
That is, \textbf{the higher the contribution of patch-wise similarity, the better re-ranking} accuracy on the challenging randomly-cropped queries.
We choose $\alpha = 0.7$ as the best and default choice for all subsequent FI experiments.
Interestingly, our proposed distance (Eq.~\ref{eq:alpha}) also yields a state-of-the-art face \emph{verification} result on MLFW \cite{wang2021mlfw} (Sec.~\ref{sec:verification}).

\subsection{Patch-wise EMD for ranking or re-ranking}
\label{sec:emd_stage_1}

Given that re-ranking using EMD at the patch-embedding space substantially improves the precision of FI compared to Stage 1 alone (\cref{tab:eff_fea}), here, we test performing such patch-wise EMD sorting at Stage 1 instead of Stage 2.

\subsec{Experiment}
That is, we test ranking images using EMD at the patch level instead of the standard cosine distance at the image level.
Performing patch-wise EMD at Stage 1 is significantly slower than our 2-stage approach, \eg, \textbf{$\sim$12 times slower} (729.20s vs. 60.97s, in total, for 13,233 queries).
That is, Sinkhorn is a slow, iterative optimization method and the EMD at Stage 2 has to sort only $k=100$ (instead of 13,233) images.
In addition, FI by comparing images patch-wise using EMD at Stage 1 yields consistently worse accuracy than our 2-stage method under all feature-weighting techniques (see \cref{tab:emd_only_vs_reranking} for details).



\section{Additional Results}
\label{sec:exper_result}

\begin{figure*}
    {
    \small
    \begin{flushleft}
    \hspace{0.5cm}
    (a) Masked (LFW)     \hspace{0.5cm} 
    (b) Sunglassess (AgeDB) \hspace{0.7cm}
    (c) Profile (CFP)    \hspace{1.0cm}
    (d) Cropped (LFW)     \hspace{0.5cm}
    (e) Adversarial (TALFW) \hfill
    \end{flushleft}
    \vspace{-0.3cm}
    }
    \centering
    \includegraphics[width=1.0\textwidth]{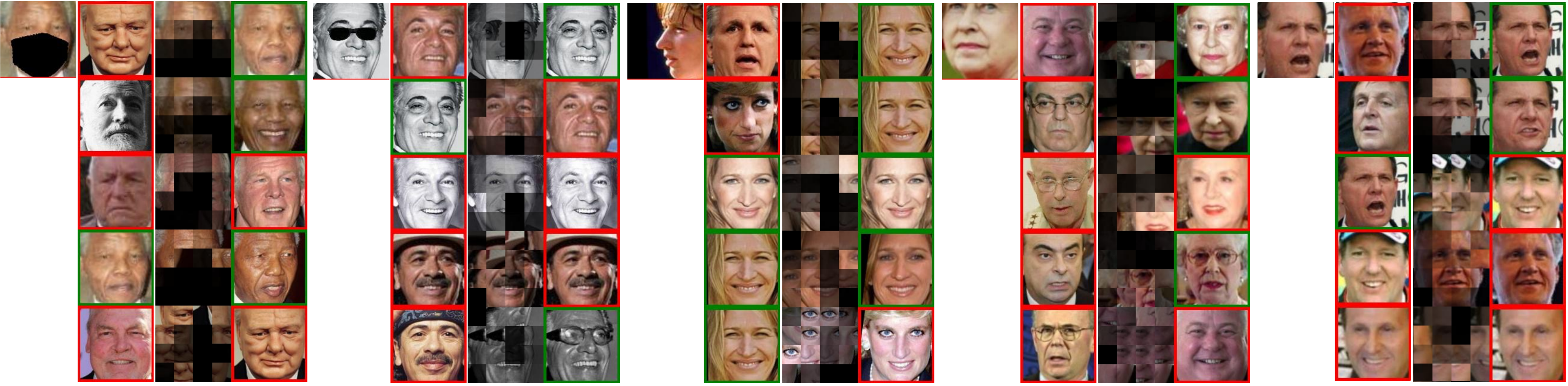}
    \scriptsize
    \begin{flushleft}
    \vspace{-0.3cm}
    \hspace{0.8cm}
    Stage 1 \hspace{0.2cm} Flow \hspace{0.1cm} Stage 2
    \hspace{0.9cm}
    Stage 1 \hspace{0.2cm} Flow \hspace{0.1cm} Stage 2
    \hspace{0.9cm}
    Stage 1 \hspace{0.2cm} Flow \hspace{0.1cm} Stage 2
    \hspace{0.9cm}
    Stage 1 \hspace{0.2cm} Flow \hspace{0.1cm} Stage 2
    \hspace{0.9cm}
    Stage 1 \hspace{0.2cm} Flow \hspace{0.1cm} Stage 2
    \hfill
    \end{flushleft}
    \vspace{-0.4cm}
    \caption{
    Figure in a similar format to that of \cref{fig:teaser}.
    Our re-ranking based on patch-wise similarity using ArcFace (4$\times$4 grid; APC) pushes more relevant gallery images higher up (here, we show top-5 results), improving face identification precision under various types of occlusions.
    The ``Flow'' visualization intuitively shows the patch-wise reconstruction of the query (top-left) given the highest-correspondence patches (\ie largest flow) from a gallery face.
    The darker a patch, the lower the flow.
    For example, despite being masked out $\sim$50\% of the face (a), Nelson Mandela can be correctly retrieved as Stage 2 finds gallery faces with similar forehead patches.
    See \cref{fig:face_vis_88} for a similar figure as the results of running our method with an 8$\times$8 grid (\ie smaller patches), which yields slightly better precision (\cref{tab:eff_fea}).
    } 
    \label{fig:face_vis}
\end{figure*}

To demonstrate the generality and effectiveness of our 2-stage FI, we take the best hyperparameter settings ($\alpha = 0.7$; APC) from the ablation studies (\cref{sec:ablation_study}) and use them for three different models (ArcFace \cite{deng2018arcface}, CosFace \cite{wang2018cosface}, and FaceNet \cite{schroff2015facenet}), which have different grid sizes.

We test the three models on five different OOD query types: 
(1) faces wearing masks or 
(2) sunglasses; 
(3) profile faces;
(4) randomly cropped faces; and
(5) adversarial faces.

\subsection{Identifying occluded faces}
\label{sec:face_occ_re}

\begin{table}[t]
\small
\hskip 0.25cm
\centering
\resizebox{7.3cm}{!}{%
\begin{tabular}{l|l|l|l|l|l}
\hline
Dataset                                          & Model                    & Method  & P@1    & RP    & M@R   \\ \hline
\multicolumn{1}{c|}{\multirow{6}{*}{\thead{CALFW \\ (Mask)}  }} & \multirow{2}{*}{\thead{ArcFace}} & ST1   & 96.81 & 53.13 & 51.70 \\ 
\multicolumn{1}{c|}{}                           &                          & Ours     & \textbf{99.92} & \textbf{57.27} & \textbf{56.33} \\ \cline{2-6} 
\multicolumn{1}{c|}{}                           & \multirow{2}{*}{CosFace} & ST1  & 98.54 & 43.46 & 41.20 \\ 
\multicolumn{1}{c|}{}                           &                          & Ours      & \textbf{99.96} & \textbf{59.85} & \textbf{58.87} \\ \cline{2-6} 
\multicolumn{1}{c|}{}                           & \multirow{2}{*}{FaceNet} & ST1   & 77.63 & 39.74 & 36.93 \\ 
\multicolumn{1}{c|}{}                           &                          & Ours     & \textbf{96.67} & \textbf{45.87} & \textbf{44.53} \\ \hline
\multirow{6}{*}{\thead{CALFW \\ (Sunglass)}}                       & \multirow{2}{*}{ArcFace} & ST1   & 51.11 & 29.38 & 26.73 \\ 
                                                &                          & Ours & \textbf{54.95} & \textbf{30.66} & \textbf{27.74} \\ \cline{2-6} 
                                                 & \multirow{2}{*}{CosFace} & ST1   & 45.20 & 25.93 & 22.78 \\ 
                                                &                          & Ours & \textbf{49.67} & \textbf{26.98} & \textbf{24.12} \\ \cline{2-6} 
                                                 & \multirow{2}{*}{FaceNet} & ST1   & 21.68 & 13.70 & 10.89 \\ 
                                                 &                          & Ours     & \textbf{25.07} & \textbf{15.04} & \textbf{12.16} \\ \hline
\multirow{6}{*}{\thead{CALFW \\ (Crop)}}                       & \multirow{2}{*}{ArcFace} & ST1   & 79.13 & 43.46 & 41.20 \\ 
                                                &                          & Ours & \textbf{92.57} & \textbf{47.17} & \textbf{45.68} \\ \cline{2-6} 
                                                 & \multirow{2}{*}{CosFace} & ST1   & 10.99 & 6.45  & 5.43  \\ 
                                                &                          & Ours      & \textbf{25.99} & \textbf{12.35} & \textbf{11.13} \\ \cline{2-6} 
                                                 & \multirow{2}{*}{FaceNet} & ST1   & 79.47 & 44.40 & 41.99 \\ 
                                                 &                          & Ours     & \textbf{85.71} & \textbf{45.91} & \textbf{43.83} \\ \hline
\multicolumn{1}{c|}{\multirow{6}{*}{\thead{AgeDB \\ (Mask)}}} & \multirow{2}{*}{ArcFace} & ST1   & 96.15 & 39.22 & 30.41 \\ 
\multicolumn{1}{c|}{}                           &                          & Ours     & \textbf{99.84} & {39.22} & \textbf{33.18} \\ \cline{2-6} 
\multicolumn{1}{c|}{}                           & \multirow{2}{*}{CosFace} & ST1   & 98.31 & 38.17 & 31.57 \\ 
\multicolumn{1}{c|}{}                           &                          & Ours     & \textbf{99.95} & \textbf{39.70} & \textbf{33.68} \\ \cline{2-6} 
\multicolumn{1}{c|}{}                           & \multirow{2}{*}{FaceNet} & ST1   & 75.99 & 22.28 & 14.95 \\ 
\multicolumn{1}{c|}{}                           &                          & Ours     & \textbf{96.53} & \textbf{24.25} & \textbf{17.49} \\ \hline
\multicolumn{1}{c|}{\multirow{6}{*}{\thead{AgeDB \\ (Sunglass)}}} & \multirow{2}{*}{ArcFace} & ST1   & 84.64 & 51.16 & 44.99 \\ 
&                          & Ours     & \textbf{87.06} & {50.40} & {44.27} \\ \cline{2-6} 
\multicolumn{1}{c|}{}                           & \multirow{2}{*}{CosFace} & ST1   & 68.93 & 34.90 & 27.30 \\ 
\multicolumn{1}{c|}{}                           &                          & Ours     & \textbf{75.97} & \textbf{35.54} & \textbf{28.12} \\ \cline{2-6} 
\multicolumn{1}{c|}{}                           & \multirow{2}{*}{FaceNet} & ST1   & 56.77 & 27.92 & 20.00 \\ 
\multicolumn{1}{c|}{}                           &                          & Ours     & \textbf{61.21} & \textbf{28.98} & \textbf{21.11} \\ \hline
\multicolumn{1}{c|}{\multirow{6}{*}{\thead{AgeDB \\ (Crop)}}} & \multirow{2}{*}{ArcFace} & ST1   & 79.92 & 32.66 & 26.19 \\ 
&                          & Ours     & \textbf{92.92} & \textbf{32.93} & \textbf{26.60} \\ \cline{2-6} 
\multicolumn{1}{c|}{}                           & \multirow{2}{*}{CosFace} & ST1   & 10.11 & 4.23 & 2.18 \\ 
&                          & Ours     & \textbf{19.58} & \textbf{4.95} & \textbf{2.76} \\ \cline{2-6} 
\multicolumn{1}{c|}{}                           & \multirow{2}{*}{FaceNet} & ST1   & 80.80 & 31.50 & 24.27 \\ 
\multicolumn{1}{c|}{}                           &                          & Ours     & \textbf{86.74} & \textbf{31.51} & \textbf{24.32} \\ \hline
\end{tabular}
}
\caption{
When the queries (from CALFW \cite{Tianyue2017calfw} and AgeDB \cite{moschoglou2017agedb}) are occluded by masks, sunglasses, or random cropping, our 2-stage method (8$\times$8 grid; APC) is substantially more robust to the Stage 1 alone baseline (ST1) with up to +13\% absolute gain (\eg P@1: 79.13 to 92.57).
The conclusions are similar for other feature-weighting methods (see \cref{tab:face_occ_calfw_full_re} and \cref{tab:face_occ_agedb_full_re}).
}
\label{tab:face_occ_re}
\end{table}

\subsec{Experiment}
We perform our 2-stage FI on three datasets: CFP \cite{Sengupta2016cfp}, CALFW \cite{Tianyue2017calfw}, and AgeDB \cite{moschoglou2017agedb}.
12,173-image CALFW and 16,488-image AgeDB have age-varying images of 4,025 and 568 identities, respectively.
CFP has 500 people, each having 14 images (10 frontal and 4 profile).


To test our models on challenging OOD queries, in CFP, we use its 2,000 profile faces in CFP as queries and its 5,000 frontal faces as the gallery.
To create OOD queries using CFP\footnote{We only apply masks and sunglasses on the frontal images of CFP.}, CALFW, and AgeDB, we automatically occlude all images with masks and sunglasses by detecting the landmarks of eyes and mouth  using \layer{dlib} and overlaying black sunglasses or a mask on the faces (see examples in \cref{fig:teaser}).
We also take these three datasets and create randomly cropped queries (as for LFW-crop in \cref{sec:ablation_study}).
For all datasets, we test identifying occluded query faces given the original, unmodified gallery.
That is, for every query, there is $\geq$ 1 matching gallery image.

\subsec{Results}
First, for all three models and all occlusion types, \ie due to masks, sunglasses, crop, and self-occlusion (profile queries in CFP), \textbf{our method consistently outperforms the traditional Stage 1 alone} approach under all three precision metrics (Tables \ref{tab:face_occ_re}, \ref{tab:face_cfp_occ_re}, \& \ref{tab:face_occ_cfp_full_re}).

Second, across all three datasets, we find the \textbf{largest improvement} that our Stage 2 provides upon the Stage 1 alone is when the queries are \textbf{randomly cropped or masked} (\cref{tab:face_occ_re}).
In some cases, the Stage 1 alone using cosine distance is not able to retrieve any relevant examples among the top-5 but our re-ranking manages to push three relevant faces into the top-5 (\cref{fig:face_vis}d). 

Third, we observe that for faces with masks or sunglasses, APC interestingly often excludes the mouth or eye regions from the fully-visible gallery faces when computing the EMD patch-wise similarity with the corresponding occluded query (\cref{fig:feature_weighting}).
The same observation can be seen in the visualizations of the most similar patch pairs, \ie highest flow, for our same 2-stage approach that uses either 4$\times$4 grids (\cref{fig:face_vis} and \cref{fig:teaser}) or 8$\times$8 grids (\cref{fig:face_vis_88}).

\subsection{Robustness to adversarial images}
\label{sec:adv_exp}

Adversarial examples pose a huge challenge and a serious security threat to computer vision systems \cite{kurakin2016adversarial,nguyen2015deep} including FI \cite{sharif2016accessorize,zhong2020towards}.
Recent research suggests that the patch representation may be the key behind ViT impressive robustness to adversarial images \cite{anonymous2022patches,shao2021adversarial,mahmood2021robustness}.
Motivated by these findings, we test our 2-stage FI on TALFW \cite{zhong2020towards} queries given an original 13,233-image LFW gallery.

\subsec{Experiment} 
TALFW contains 4,069 LFW images perturbed adversarially to cause face verifiers to mislabel \cite{zhong2020towards}.

\subsec{Results} 
Over the entire TALFW query set, we find our re-ranking to consistently outperform the Stage 1 alone under all three metrics (\cref{tab:adv_re}).
Interestingly, the improvement (of $\sim$2 to 4 points under P@1 for three models) is larger than when tested on the original LFW queries (around 0.12 in \cref{tab:eff_fea}a), verifying our patch-based re-ranking robustness when queries are perturbed with very small noise.
That is, our approach can improve FI precision when the perturbation size is either small (adversarial) or large (\eg masks).

\begin{table}[t]
\small
\centering
\resizebox{8.3cm}{!}{%
\begin{tabular}{l|l|l|l|l|l}
\hline
Dataset                                          & Model                    & Method  & P@1    & RP    & M@R   \\ \hline
\multicolumn{1}{c|}{\multirow{6}{*}{\thead{TALFW ~\cite{zhong2020towards} \\ vs. \\ LFW ~\cite{yi2014learning} }}} & \multirow{2}{*}{ArcFace} & ST1  & 93.49 & 81.04 & 80.35 \\ 
\multicolumn{1}{c|}{}                           &                          & Ours     & \textbf{96.64} & \textbf{82.72} & \textbf{82.10} \\ \cline{2-6} 
\multicolumn{1}{c|}{}                           & \multirow{2}{*}{CosFace} & ST1  & 96.49 & 83.57 & 82.99 \\ 
\multicolumn{1}{c|}{}                           &                          & Ours     & \textbf{99.07} & \textbf{85.48} & \textbf{85.03} \\ \cline{2-6} 
\multicolumn{1}{c|}{}                           & \multirow{2}{*}{FaceNet} & ST1  & 95.33 & 79.24 & 78.19 \\ 
\multicolumn{1}{c|}{}                           &                          & Ours     & \textbf{97.26} & \textbf{80.33} & \textbf{79.39} \\ \hline 
\end{tabular}
}
\caption{
Our re-ranking (8$\times$8 grid; APC) consistently improves the precision over Stage 1 alone (ST1) when identifying adversarial TALFW \cite{zhong2020towards} images given an in-distribution LFW \cite{yi2014learning} gallery.
The conclusions also carry over to other feature-weighting methods (more results in \cref{tab:face_occ_talfw_full_re}).
}
\label{tab:adv_re}
\end{table}


\subsection{Re-ranking rivals finetuning on masked images}
\label{sec:com_finetune}

While our approach does not involve re-training, a common technique for improving FI robustness to occlusion is data augmentation, \ie re-train the models on occluded data in addition to the original data.
Here, we compare our method with data augmentation on masked images.

\subsec{Experiment}
To generate augmented, masked images, we follow \cite{Anwar2020MaskedFR} to overlay various types of masks on CASIA images to generate $\sim$415K masked images.
We add these images to the original CASIA training set, resulting in a total of $\sim$907K images (10,575 identities).
We finetune ArcFace on this dataset with the same original hyperparameters \cite{arcfacePyTorch} (see \cref{sec:finetuning_hyperparameters}).
We train three models and report the mean and standard deviation (\cref{tab:comp_finetune}).

For a fair comparison, we evaluate the finetuned models and our no-training approach on the MLFW dataset \cite{wang2021mlfw}, instead of our self-created masked datasets.
That is, the query set has 11,959 MLFW masked-face images and the gallery is the entire 13,233-image LFW.

\subsec{Results}
First, we find that finetuning ArcFace improves its accuracy in FI under Stage 1 alone (\cref{tab:comp_finetune}; 39.79 vs. 41.64).
Yet, our 2-stage approach still substantially outperforms Stage 1 alone, both when using the original and the finetuned ArcFace (\cref{tab:comp_finetune}; 48.23 vs. 41.64).
Interestingly, we also test using the finetuned model in our DeepFace-EMD framework and finds it to approach closely the best no-training result (46.21 vs. 48.23).

\begin{table}[h]
\small
\centering
\resizebox{8.5cm}{!}{%
\begin{tabular}{l|l|l|l|l} 
\hline
\thead{ArcFace} &Method & P@1 & RP & M@R \\
\hline
\multirow{2}{*}{Pre-trained} & (a)\quad ST1 & 39.79 & 35.10 & 33.32\\ 
& (b)\quad Ours & \textbf{48.23 }& \textbf{41.43} & \textbf{39.71}  \\
\hline
\multirow{2}{*}{Finetuned} & (c)\quad ST1 & $41.64 \pm$ \small 0.16\par  & $34.67 \pm$ 0.24 & $32.66 \pm$ 0.25 \\ 
& (d)\quad Ours & 46.21 $\pm$ 0.27 & 38.65 $\pm$ 0.26 & $36.73 \pm$ 0.26 \\ 
\hline 
\end{tabular}
}
\caption{
Our 2-stage approach (b) using ArcFace (8$\times$8 grid; APC) substantially outperforms Stage 1 alone (a) on identifying masked images of MLFW given the unmasked gallery of LFW.
Interestingly, our method (b) also outperforms Stage 1 alone when ArcFace has been finetuned on masked images (c).
In (c), we report the mean and std over three finetuned models.
}
\label{tab:comp_finetune}
\vspace{-1em}
\end{table}



\section{Related Work}
\label{sec:related_work}

\paragraph{Face Identification under Occlusion}

Partial occlusion presents a significant, ill-posed challenge to face identification as the AI has to rely only on incomplete or noisy facial features to make decisions \cite{qiu2021end2end}.
Most prior methods propose to improve FI robustness by augmenting the training set of deep feature extractors with partially-occluded faces \cite{trigueros2018enhancing,wang2021mlfw,osherov2017increasing,xu2020improving,guo2018face,xu2020improving}.
Training on augmented, occluded data encourages models to rely more on local, discriminative facial features \cite{osherov2017increasing}; however, does not prevent FI models from misbehaving on new OOD occlusion types, especially under adversarial scenarios \cite{sharif2016accessorize}.
In contrast, our approach (1) does not require re-training or data augmentation; and (2) harnesses both image-level features (stage 1) and local, patch-level features (stage 2) for FI.

A common alternative is to learn to generate a spatial feature mask \cite{Song2019Occ,wan2017occlusion,qiu2021end2end,min2011improving} or an attention map \cite{xu2020improving} to \emph{exclude} the occluded (\ie uninformative or noisy) regions in the input image from the face matching process.
Motivated by these works, we tested five methods for inferring the importance of each image patch (\cref{sec:ablation_study}) for EMD computation.
Early works used hand-crafted features and obtained limited accuracy \cite{min2011improving,oh2008occlusion,li2005nonparametric}.
In contrast, the latter attempts took advantage of deep architectures but requires a separate occlusion detector \cite{Song2019Occ} or a masking subnetwork in a custom architecture trained end-to-end \cite{qiu2021end2end,wan2017occlusion}.
In contrast, we leverage directly the pre-trained state-of-the-art image embeddings (of ArcFace, CosFace, \& FaceNet) and EMD to exclude the occluded regions from an input image \emph{without any} architectural modifications or re-training.

Another approach is to predict occluded pixels and then perform FI on the recovered images \cite{zhao2017robust,wright2008robust,zhou2009face,yang2011robust,he2011regularized,li2013structured}.
Yet, how to recover a non-occluded face while preserving true identity remains a challenge to state-of-the-art GAN-based de-occlusion methods \cite{dong2020occlusion,cai2020semi,ge2020occluded}.

\subsec{Re-ranking in Face Identification}
Re-ranking is a popular 2-stage method for refining image retrieval results \cite{zhang2020understanding} in many domains, \eg person re-identification \cite{sarfraz2018pose}, localization \cite{shen2012object}, or web image search \cite{cui2008real}.
In FI, Zhou et. al. \cite{zhou2020AsArcFaceAA} used hand-crafted patch-level features to encode an image for ranking and then used multiple reference images in the database to re-rank each top-$k$ candidate.
The social context between two identities has also been found to be useful in re-ranking photo-tagging results \cite{bharadwaj2014aiding}.
Swearingen et. al. \cite{swearingen2021lookalike} found that harnessing an external ``disambiguator'' network trained to separate a query from lookalikes is an effective re-ranking method.
In contrast to the prior work, we do not use extra images \cite{zhou2020AsArcFaceAA} or external knowledge \cite{bharadwaj2014aiding}.
Compared to face re-ranking \cite{Dong2020OpensetFI, Peng2019TIP}, our method is the first re-rank candidates based on a pair-wise similarity score computed from both the image-level and patch-level similarity computed off of state-of-the-art deep facial features.

\subsec{EMD for Image Retrieval}
While EMD is a well-known metric in image retrieval \cite{rubner2000earth}, its applications on \emph{deep} convolutional features of images have been relatively under-explored.
Zhang et al.~\cite{Zhang_2020_CVPR, zhang2020deepemdv2} recently found that classifying fine-grained images (of dogs, birds, and cars) by comparing them patch-wise using EMD in a deep feature space improves few-shot fine-grained classification accuracy.
Yet, their success has been limited to \emph{few}-shot, 5-way and 10-way classification with smaller networks (ResNet-12 \cite{he2016identity}).
In contrast, here, we demonstrate a substantial improvement in FI using EMD without re-training the feature extractors.

Concurrent to our work, Zhao et al. \cite{zhao2021towards} proposes DIML, which exhibits consistent improvement of $\sim$2--3\% in image retrieval on images of birds, cars, and products by using the sum of cosine distance and EMD as a ``structural similarity'' score for ranking.
They found that CC is more effective than assigning uniform weights to image patches \cite{zhao2017robust}.
Interestingly, via a rigorous study into different feature-weighting techniques, we find novel insights specific for FI: Uniform weighting is more effective than CC.
Unlike prior EMD works \cite{zhao2021towards,Zhang_2020_CVPR,zhang2020deepemdv2,wang2012supervised}, ours is \textbf{the first to} show the significant effectiveness of EMD on (1) occluded and adversarial OOD images; and (2) on face identification.

\section{Discussion and Conclusion}
\label{sec:con}

\subsec{Limitations}
Solving patch-wise EMD via Sinkhorn is slow, which may prohibit it from being used to sort a much larger image sets (see run-time reports in \cref{tab:emd_only_vs_reranking}).
Furthermore, here, we used EMD on two distributions of equal weights; however, the algorithm can be used for unequal-weight cases \cite{rubner2000earth,cohen1999finding}, which may be beneficial for handling occlusions.
While substantially improving FI accuracy under the four occlusion types (\ie, masks, sunglasses, random crops, and adversarial images), re-ranking is only marginally better than Stage 1 alone on ID and profile faces, which is interesting to understand deeper in future research.

Instead of using pre-trained models, it might be interesting to re-train new models explicitly on patch-wise correspondence tasks, which may yield better patch embeddings for our re-ranking.
In sum, we propose DeepFace-EMD, a 2-stage approach for comparing images hierarchically: First at the image level and then at the patch level.
DeepFace-EMD shows impressive robustness to occluded and adversarial faces and can be easily integrated into existing FI systems in the wild.



\subsec{Acknowledgement}
We thank Qi Li, Peijie Chen, and Giang Nguyen for their feedback on manuscript.
We also thank Chi Zhang, Wenliang Zhao, Chengrui Wang for releasing their DeepEMD, DIML, and MLFW code, respectively.
AN was supported by the NSF Grant No. 1850117 and a donation from NaphCare Foundation.

{\small
\bibliographystyle{ieee_fullname}
\bibliography{references}
}

\newpage
\clearpage

\renewcommand{\thesection}{S\arabic{section}}
\renewcommand{\thesubsection}{\thesection.\arabic{subsection}}

\newcommand{\beginsupplementary}{%
            \setcounter{table}{0}
    \renewcommand{\thetable}{S\arabic{table}}%
            \setcounter{figure}{0}
    \renewcommand{\thefigure}{S\arabic{figure}}%
    \setcounter{section}{0}
}
\newcommand{\suptitle}{Appendix for:\\\papertitle}

\newcommand{\toptitlebar}{
    \hrule height 4pt
    \vskip 0.25in
    \vskip -\parskip%
}
\newcommand{\bottomtitlebar}{
    \vskip 0.29in
    \vskip -\parskip%
    \hrule height 1pt
    \vskip 0.09in%
}

\beginsupplementary%

\newcommand{\maketitlesupp}{
    \onecolumn
    \begin{@twocolumnfalse}
        \null%
        \vskip .375in
        \begin{center}
            {\Large \bf \suptitle\par}
            \vspace*{24pt}
            {
                \large
                \lineskip=.5em
                \par
            }
            \vskip .5em
            \vspace*{12pt}
        \end{center}
    \end{@twocolumnfalse}
}

\maketitlesupp

\section{Pre-trained models}
\label{sec:supp_pretrained_models}

\paragraph{Sources} We downloaded the three pre-trained PyTorch models of ArcFace, FaceNet, and CosFace from:

\begin{itemize}
    \item ArcFace \cite{deng2018arcface}: \url{https://github.com/ronghuaiyang/arcface-pytorch}
    
    \item FaceNet \cite{schroff2015facenet}: \url{https://github.com/timesler/facenet-pytorch}
    
    \item CosFace \cite{wang2018cosface}: \url{https://github.com/MuggleWang/CosFace_pytorch}
\end{itemize}

These ArcFace, FaceNet, and CosFace models were trained on dataset CASIA Webface \cite{yi2014learning}, VGGFace2 ~\cite{Cao2018VGGFace2AD}, and CASIA Webface ~\cite{yi2014learning}, respectively.

\paragraph{Architectures} The network architectures are provided here:

\begin{itemize}
    \item ArcFace: \url{https://github.com/ronghuaiyang/arcface-pytorch/blob/master/models/resnet.py}\item FaceNet: \url{https://github.com/timesler/facenet-pytorch/blob/master/models/inception_resnet_v1.py}
    \item CosFace: \url{https://github.com/MuggleWang/CosFace_pytorch/blob/master/net.py#L19}
\end{itemize}

\paragraph{Image-level embeddings for Ranking} We use these layers to extract the image embeddings for stage 1, \ie, ranking images based on the cosine similarity between each pair of (query image, gallery image).

\begin{itemize}
    \item Arcface: layer~\layer{bn5} (see \href{https://github.com/ronghuaiyang/arcface-pytorch/blob/master/models/resnet.py#L177}{code}), which is the 512-output, last BatchNorm linear layer of ArcFace (a modified ResNet-18 \cite{he2016identity}).
    \item FaceNet: layer~\layer{last\_bn} (see \href{https://github.com/timesler/facenet-pytorch/blob/master/models/inception_resnet_v1.py#L258}{code}), which is the 512-output, last BatchNorm linear layer of FaceNet (an Inception-ResNet-v1 \cite{szegedy2017inception}).
    \item CosFace: layer~\layer{fc} (see \href{https://github.com/MuggleWang/CosFace_pytorch/blob/master/net.py#L37}{code}), which is the 512-output, last linear layer of the 20-layer SphereFace architecture \cite{Liu_2017_CVPR}.
\end{itemize}


\paragraph{Patch-level embeddings for Re-ranking}

We use the following layers to extract the spatial feature maps (\ie embeddings $\{q_i\}$) for the patches:

\begin{itemize}
    \item ArcFace: layer \layer{dropout} (see \href{https://github.com/ronghuaiyang/arcface-pytorch/blob/master/models/resnet.py#L175}{code}). Spatial dimension: $8 \times 8$.
    
    \item FaceNet: layer \layer{block8} (see \href{https://github.com/timesler/facenet-pytorch/blob/master/models/inception_resnet_v1.py#L254}{code}) Spatial dimension: $3 \times 3$.
    
    \item CosFace: layer \layer{layer4} (see \href{https://github.com/MuggleWang/CosFace_pytorch/blob/master/net.py#L36}{code}). Spatial dimension: $6\times 7$.
\end{itemize}

\section{Finetuning hyperparameters}
\label{sec:finetuning_hyperparameters}
\begin{figure*}
    \centering
    \includegraphics[width=0.7\textwidth]{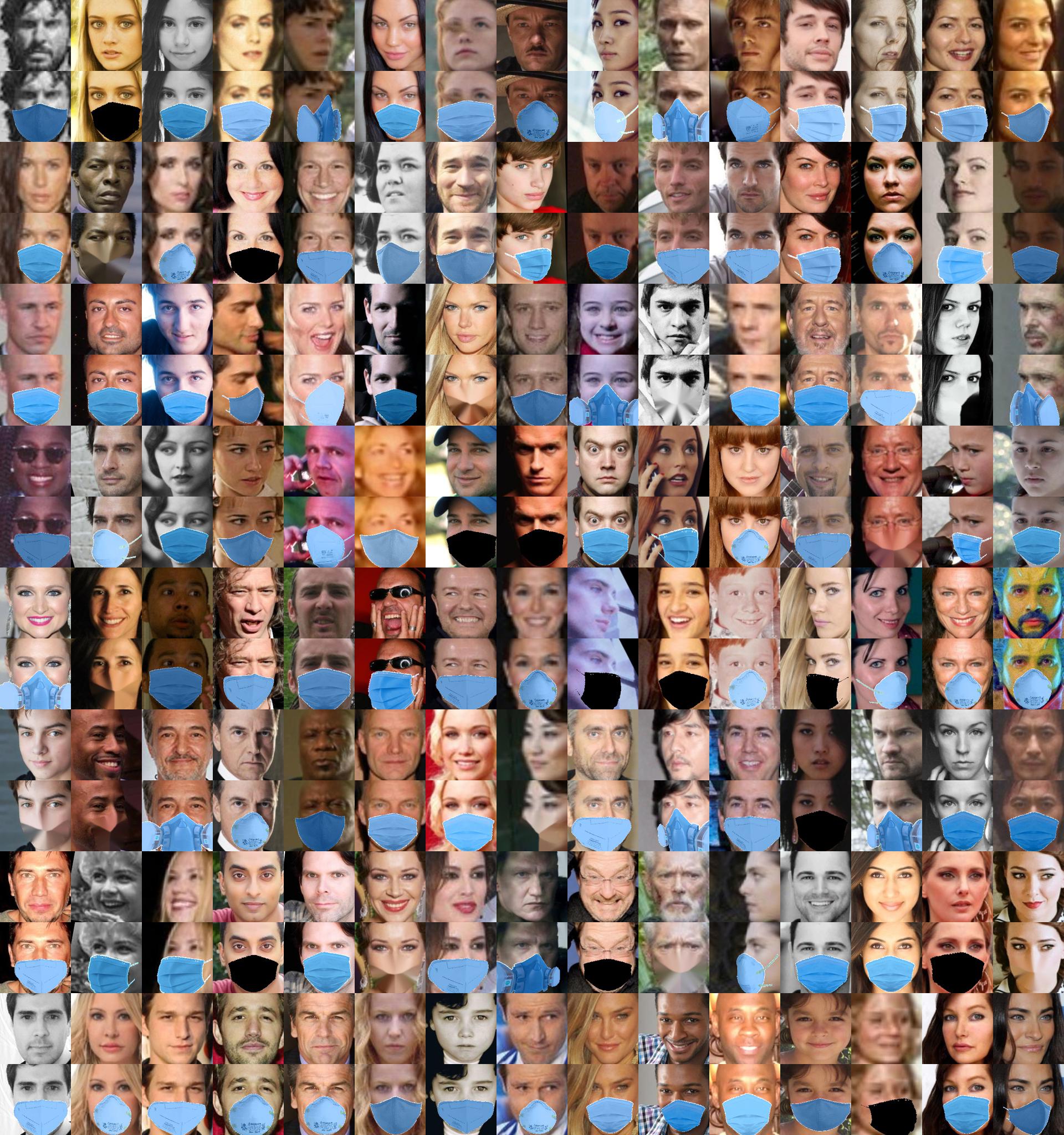}
    \scriptsize
    \caption{Our CASIA dataset augmented with masked images (generated following the method by \cite{Anwar2020MaskedFR}) for fine-tuning ArcFace.} 
    \label{fig:face_casia_mask_vis}
\end{figure*}

We describe here the hyperparameters used for finetuning ArcFace on our CASIA dataset augmented with masked images (see \cref{fig:face_casia_mask_vis} for some samples).
\begin{itemize}
    \item Training on $907,459$ facial images (masks and non-masks).
    \item Number of epochs is $12$.
    \item Optimizer: SGD.
    \item Weight decay: $5e^{-4}$
    \item Learning rate: $0.001$
    \item Margin: $m=0.5$
    \item Feature scale: $s=30.0$
\end{itemize}

See details in the published code base: \href{https://github.com/ronghuaiyang/arcface-pytorch/blob/master/train.py}{code}

\section{Flow visualization}
\label{sec:flow_vis_description}

\begin{figure}[h]
    {
    \begin{flushleft}
        \hspace{1.5cm}
        SC \hspace{3.5cm} 
        APC \hspace{4cm}
        LMK  \hspace{3.5cm}
        Uniform
    \end{flushleft}
    
    }
\centering
    \includegraphics[width=1.0\textwidth]{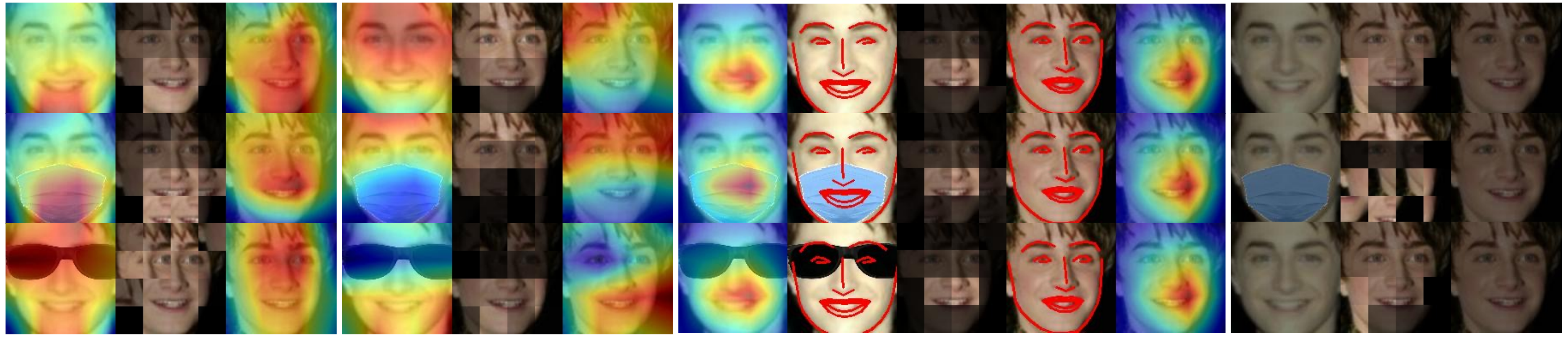}
    \caption{
    Given a pair of images, after the features are weighted (heatmaps; red corresponds to 1 and blue corresponds to 0 importance weight), EMD computes an optimal matching or ``transport'' plan.
    The middle flow image shows the one-to-one correspondence following the format in \cite{zhang2020deepemdv2} (see also description in \cref{sec:flow_vis_description}).
    That is, intuitively, the flow visualization shows the reconstruction of the left image, using the nearest patches (i.e. highest flow) from the right image.
    Here, we use ArcFace and a 4$\times$ patch size (\ie computing the EMD between two sets of 16 patch-embeddings).
    Darker patches correspond to smaller flow values.
    How EMD computes facial patch-wise similarity differs across different feature weighting techniques (SC, APC, LMK, and Uniform).
    %
    }
    \label{fig:heatmap_flows_single}
\end{figure}

We use the same visualization technique as in DeepEMD to generate the flow visualization showing the correspondence between two images (see the flow visualization in \cref{fig:teaser} or \cref{fig:heatmap_flows_single}).
Given a pair of embeddings from query and gallery images, EMD computes the optimal flows (see \cref{eq:emd_opt} for details). 
That is, given a 8$\times$8 grid, a given patch embedding $q_i$ in the query has 64 flow values $\{f_{ij}\}$ where $j \in \{1, 2, ..., 64\}$.
In the location of patch $q_i$ in the query image, we show the corresponding highest-flow patch $g_k$, \ie $k$ is the index of the gallery patch of highest flow $f_{i,k} = \max ( f_{i,1}, f_{i,2}, ..., f_{i,64} )$.
For displaying, we normalize a flow value $f_{i,k}$ over all 64 flow values (each for a patch $i \in \{1, 2, ..., 64\}$) via:

\begin{align}
    f = \frac{f - \min(f)}{\max(f) - \min(f)}
\end{align}

\begin{figure*}
    {
    \small
    \begin{flushleft}
    \hspace{1cm}
    (a) LFW \hspace{2.5cm}
    (b) Masked    \hspace{2.5cm}
    (c) Sunglasses (LFW)     \hspace{2.5cm}
    (d) Profile (CFP) \hfill
    \end{flushleft}
    \vspace{-0.3cm}
    }
    \centering
    \includegraphics[width=1.0\textwidth]{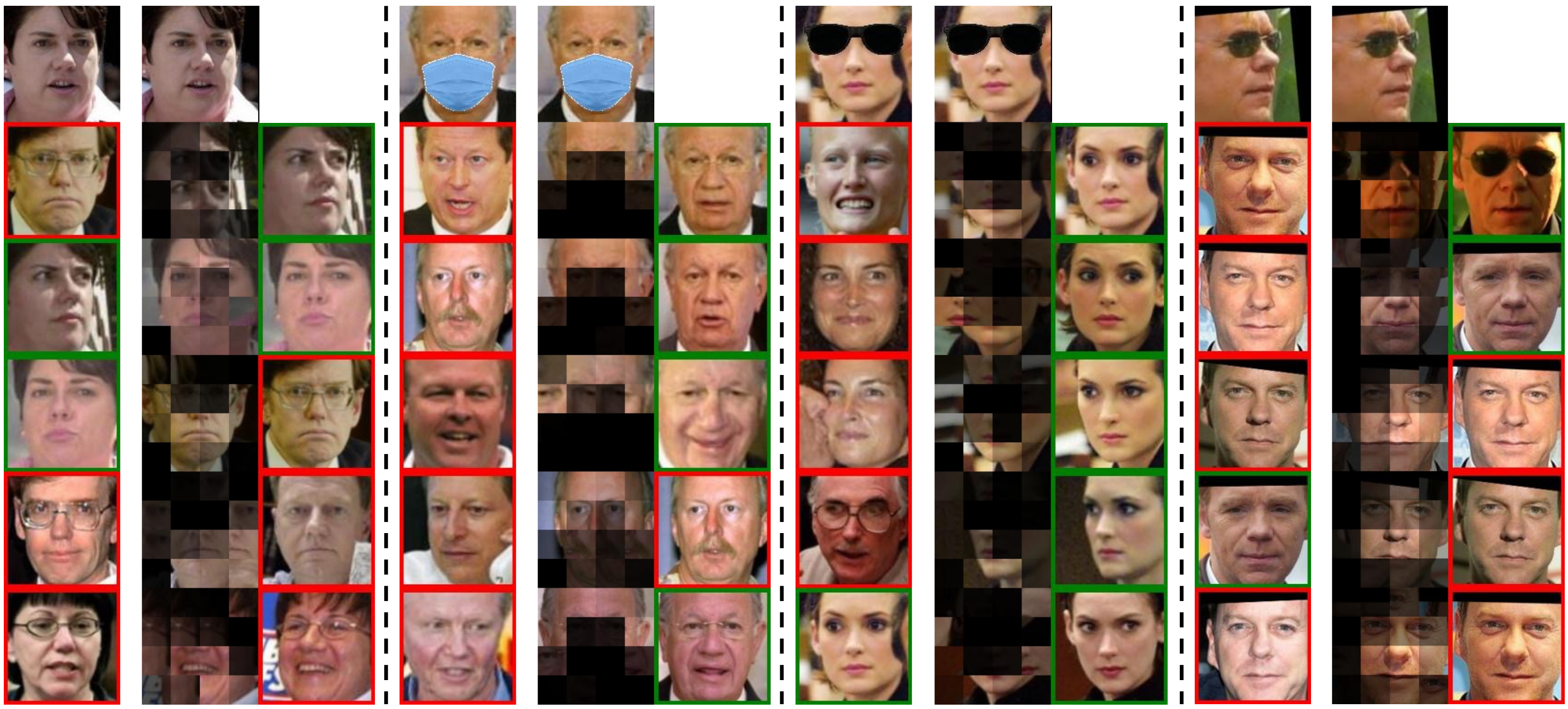}
    \scriptsize
    \vspace{-0.5cm}
    \begin{flushleft}
    \hspace{0.3cm}
    Stage 1 \hspace{0.7cm} Flow \hspace{0.5cm} Stage 2
    \hspace{0.8cm}
    Stage 1 \hspace{0.7cm} Flow \hspace{0.5cm} Stage 2
    \hspace{0.8cm}
    Stage 1 \hspace{0.7cm} Flow \hspace{0.5cm} Stage 2
    \hspace{1.0cm}
    Stage 1 \hspace{0.7cm} Flow \hspace{0.5cm} Stage 2
    \hfill
    \end{flushleft}
    \caption{
    Traditional face identification ranks gallery images based on their cosine distance with the query (top row) at the image-level embedding, which yields large errors upon out-of-distribution changes in the input (\eg masks or sunglasses; b--d).
    We find that re-ranking the top-$k$ shortlisted faces from Stage 1 (leftmost column) using their patch-wise EMD similarity w.r.t. the query substantially improves the precision (Stage 2) on challenging cases (b--d).
    The ``Flow'' visualization (of $4 \times 4$) intuitively shows the patch-wise reconstruction of the query face using the most similar patches (\ie highest flow) from the retrieved face.
    } 
    
    \label{fig:face_vis_44}
\end{figure*}

\begin{figure*}
    {
    \small
    \begin{flushleft}
    \hspace{1cm}
    (a) LFW \hspace{2.5cm}
    (b) Masked (LFW)    \hspace{2.5cm}
    (c) Sunglasses (LFW)    \hspace{2.5cm}
    (d) Profile (CFP) \hfill
    \end{flushleft}
    \vspace{-0.3cm}
    }
    \centering
    \includegraphics[width=1.0\textwidth]{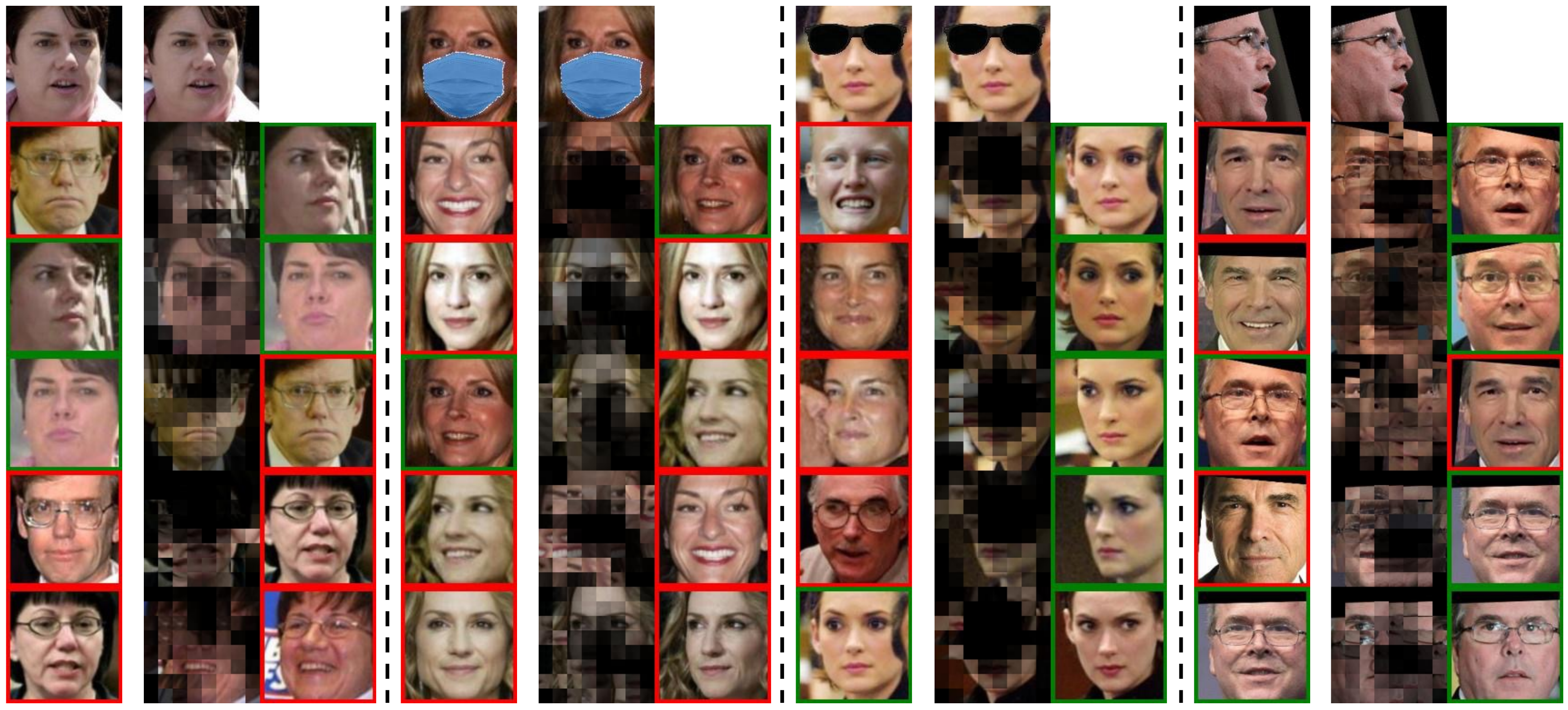}
    \scriptsize
    \vspace{-0.5cm}
    \begin{flushleft}
    \hspace{0.3cm}
    Stage 1 \hspace{0.7cm} Flow \hspace{0.5cm} Stage 2
    \hspace{0.8cm}
    Stage 1 \hspace{0.7cm} Flow \hspace{0.5cm} Stage 2
    \hspace{0.8cm}
    Stage 1 \hspace{0.7cm} Flow \hspace{0.5cm} Stage 2
    \hspace{1.0cm}
    Stage 1 \hspace{0.7cm} Flow \hspace{0.5cm} Stage 2
    \hfill
    \end{flushleft}
    \caption{
    Traditional face identification ranks gallery images based on their cosine distance with the query (top row) at the image-level embedding, which yields large errors upon out-of-distribution changes in the input (\eg masks or sunglasses; b--d).
    We find that re-ranking the top-$k$ shortlisted faces from Stage 1 (leftmost column) using their patch-wise EMD similarity w.r.t. the query substantially improves the precision (Stage 2) on challenging cases (b--d).
    The ``flow'' visualization (of $8 \times 8$) intuitively shows the patch-wise reconstruction of the query face using the most similar patches (\ie highest flow) from the retrieved face.
    } 
    \label{fig:face_vis_88}
\end{figure*}

See \cref{fig:face_vis_44}, \cref{fig:face_vis_88}, and \cref{fig:face_vis} for example flow visualizations.      
\begin{figure}[h]
\centering
    \includegraphics[width=0.7\textwidth]{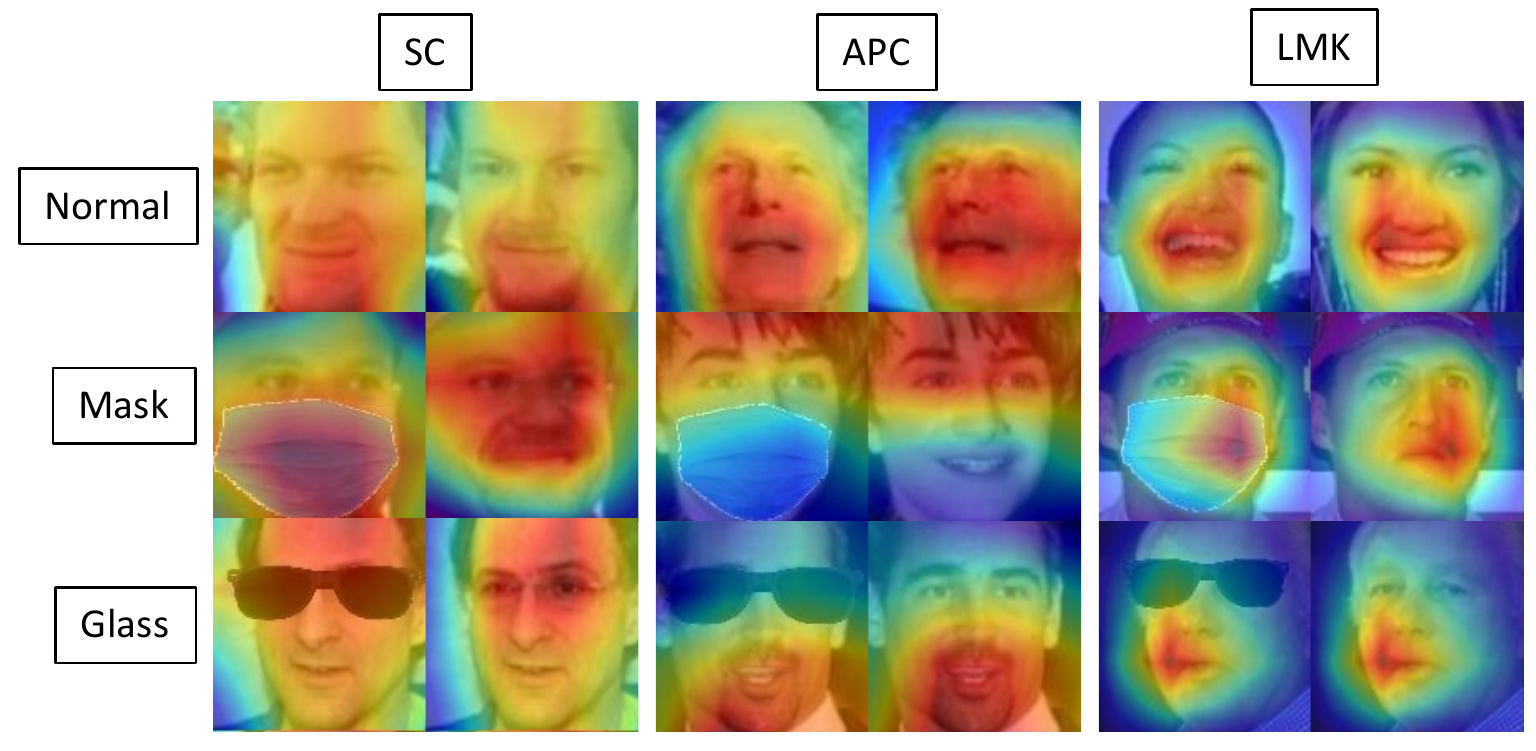}
    \caption{
        The feature-weighting heatmaps using SC, APC, and LMK for random pairs of faces across three input types (normal faces, and faces with masks and sunglasses).
        Here, we use ArcFace \cite{deng2018arcface} and an 4$\times$4 grid (average pooling result from 8$\times$8).
        SC heatmaps often cover the entire face including the occluded region.
        APC tend to assign low importance to occlusion and the corresponding region in the unoccluded image (see blue areas in APC).
        LMK results in a heatmap that covers the middle area of a face.
        Best view in color.
    }
    \label{fig:heatmap}
\end{figure}


\begin{table}[]
\centering
\begin{tabular}{l|cc|r|c|c|c}
\cline{1-7}
ArcFace                   & \multicolumn{2}{c|}{Method}                                & Time (s)        & P@1            & RP             & MAP@R           \\ \cline{1-7}
\multirow{8}{*}{(a) LFW}      & \multicolumn{1}{c|}{\multirow{2}{*}{APC}}     & EMD at Stage 1   & 268.96          & 83.35          & 76.97          & 73.81           \\ 
                          & \multicolumn{1}{c|}{}                         & Ours & \textbf{60.03}  & \textbf{98.60} & \textbf{78.63} & \textbf{78.22}            \\ \cline{2-7}
                          & \multicolumn{1}{c|}{\multirow{2}{*}{SC}}      & EMD at Stage 1   & 196.50          & 97.85          & 77.92          & 77.29                              \\ 
                          & \multicolumn{1}{c|}{}                         & Ours & \textbf{77.32} & \textbf{98.66} & \textbf{78.74} & \textbf{78.35}             \\ \cline{2-7}
                          & \multicolumn{1}{c|}{\multirow{2}{*}{Uniform}} & EMD at Stage 1   & 191.47          & 97.85          & 77.91          & 77.29           \\ 
                          & \multicolumn{1}{c|}{}                         & Ours & \textbf{77.79} & \textbf{98.66} & \textbf{78.73} & \textbf{78.35} \\ \cline{2-7}
                          & \multicolumn{1}{c|}{\multirow{2}{*}{LMK}}     & EMD at Stage 1   & 178.67          & 98.13          & 78.18          & 77.61           \\ 
                          & \multicolumn{1}{c|}{}                         & Ours & \textbf{77.79} & \textbf{98.66} & \textbf{78.73} & \textbf{78.35}  \\ \cline{1-7}
                          
\multirow{6}{*}{\thead{(b) LFW-crop \\ vs. \\ LFW}} & \multicolumn{1}{c|}{\multirow{2}{*}{APC}}     & EMD at Stage 1   & 729.20          & 55.53          & 44.06          & 38.57\\ 
                          & \multicolumn{1}{c|}{}                         & Ours & \textbf{60.97}  & \textbf{96.10} & \textbf{76.58} & \textbf{74.56}             \\ \cline{2-7}
                          & \multicolumn{1}{c|}{\multirow{2}{*}{SC}}      & EMD at Stage 1   & 266.74          & 98.57          & 76.20          & 74.30            \\ 
                          & \multicolumn{1}{c|}{}                         & Ours & \textbf{60.39}  & {96.19} & \textbf{78.05} & \textbf{76.20}   \\ \cline{2-7}
                          & \multicolumn{1}{c|}{\multirow{2}{*}{Uniform}} & EMD at Stage 1   & 259.84          & \textbf{98.62}          & 76.19          & 74.28          \\ 
                          & \multicolumn{1}{c|}{}                         & Ours & \textbf{61.81}  & {96.26} & \textbf{78.08} & \textbf{76.25}  \\ \cline{1-7}
\end{tabular}
\caption{Comparison of performing patch-wise EMD ranking at Stage 1 vs. our proposed 2-stage FI approach (\ie cosine similarity ranking in Stage 1 and patch-wise EMD re-ranking in Stage 2). 
In both cases, EMD uses 8$\times$8 patches.
EMD at Stage 1 is the method of using EMD to rank images directly (instead of the regular cosine similarity) and there is no Stage 2 (re-ranking). 
For our method, we choose the same setup of $\alpha = 0.7$. 
Our 2-stage approach does not only outperform using EMD at Stage 1 but is also $\sim$2-4 $\times$ faster.
The run time is the total for all \textbf{13,214 queries} for both (a) and (b).
The result supports our choice of performing EMD in Stage 2 instead of Stage 1.
}
\label{tab:emd_only_vs_reranking}
\end{table}

\begin{figure*}[h!]
    \centering 
    \begin{subfigure}{0.32\textwidth}
        \includegraphics[width=\linewidth]{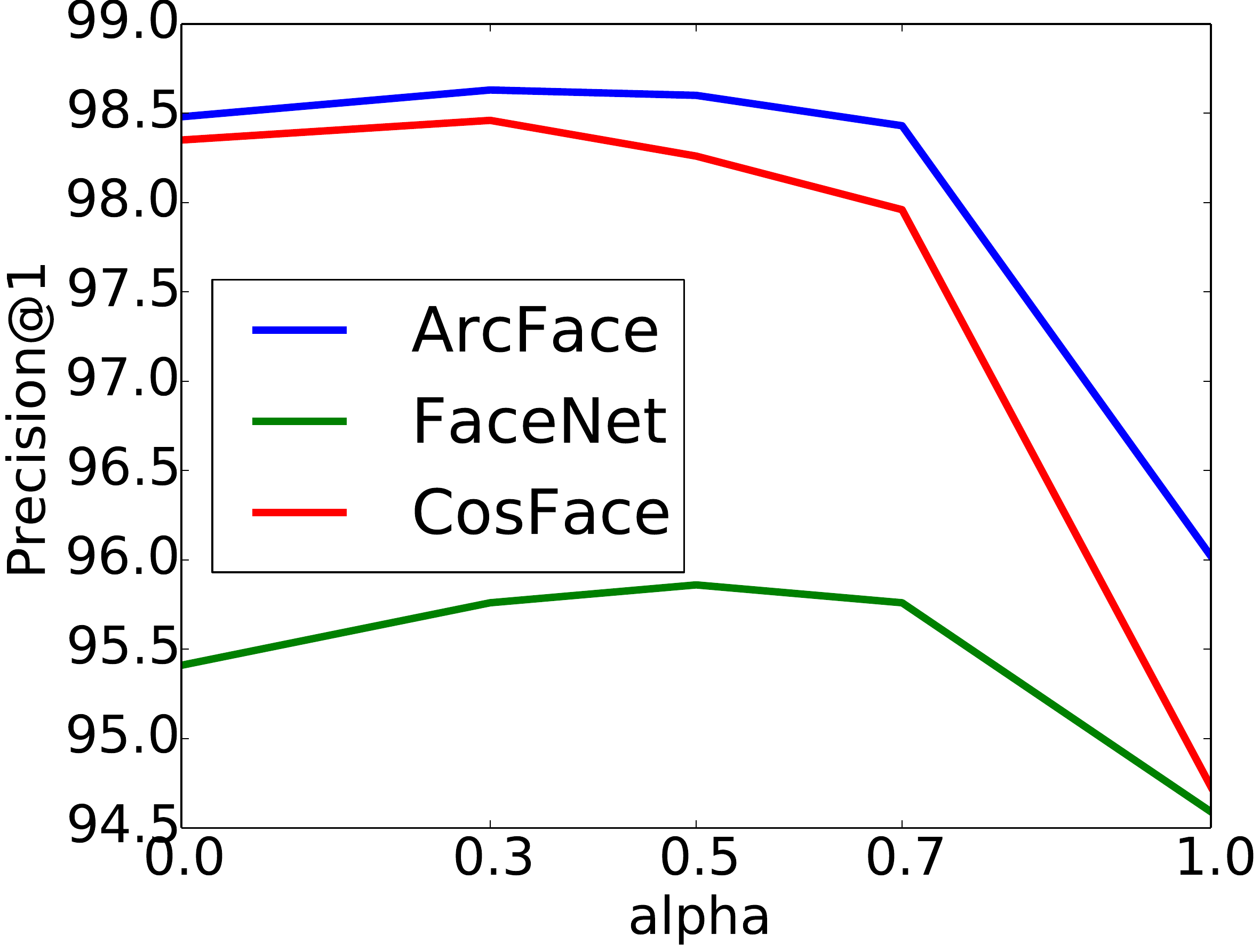}
        \caption{APC (LFW)}
        \label{fig:apc}
    \end{subfigure}\bigskip
    \begin{subfigure}{0.32\textwidth}
        \includegraphics[width=\linewidth]{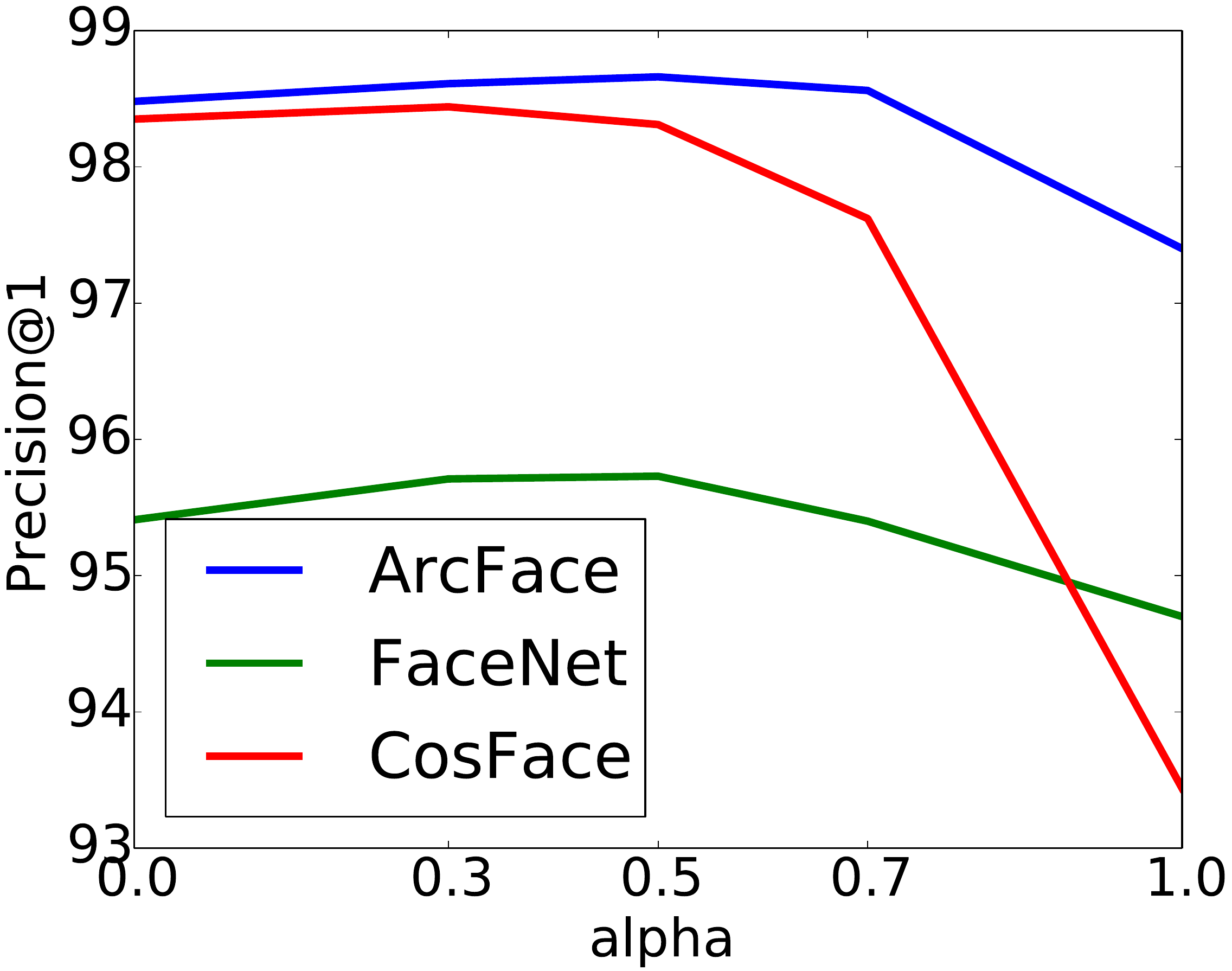}
        \caption{Uniform (LFW)}
        \label{fig:uniform}
    \end{subfigure} \bigskip 
    \begin{subfigure}{0.32\textwidth}
        \includegraphics[width=\linewidth]{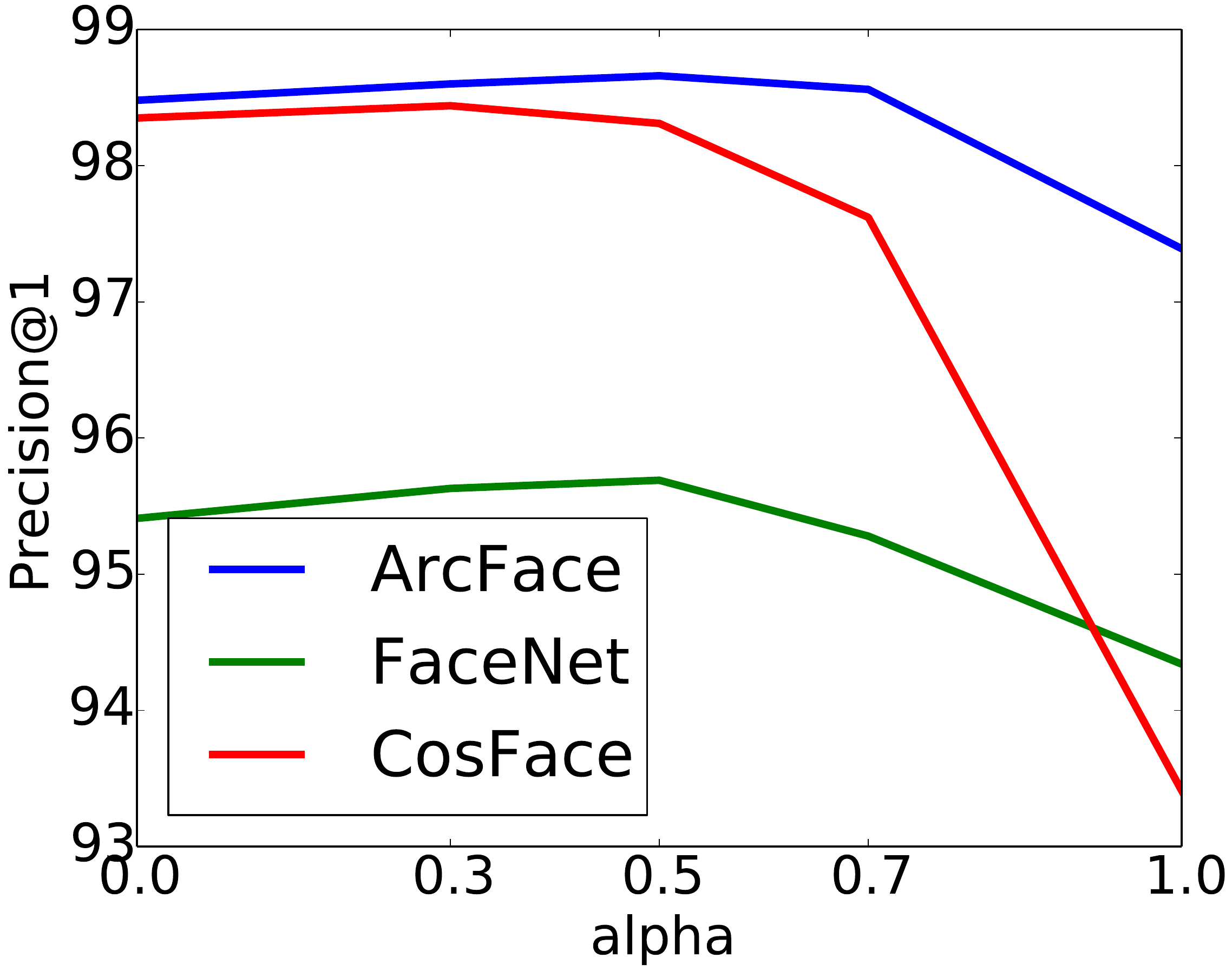}
        \caption{SC (LFW)}
        \label{fig:sc}
    \end{subfigure}

    \vspace*{-8mm}
    \begin{subfigure}{0.32\textwidth}
        \includegraphics[width=\linewidth]{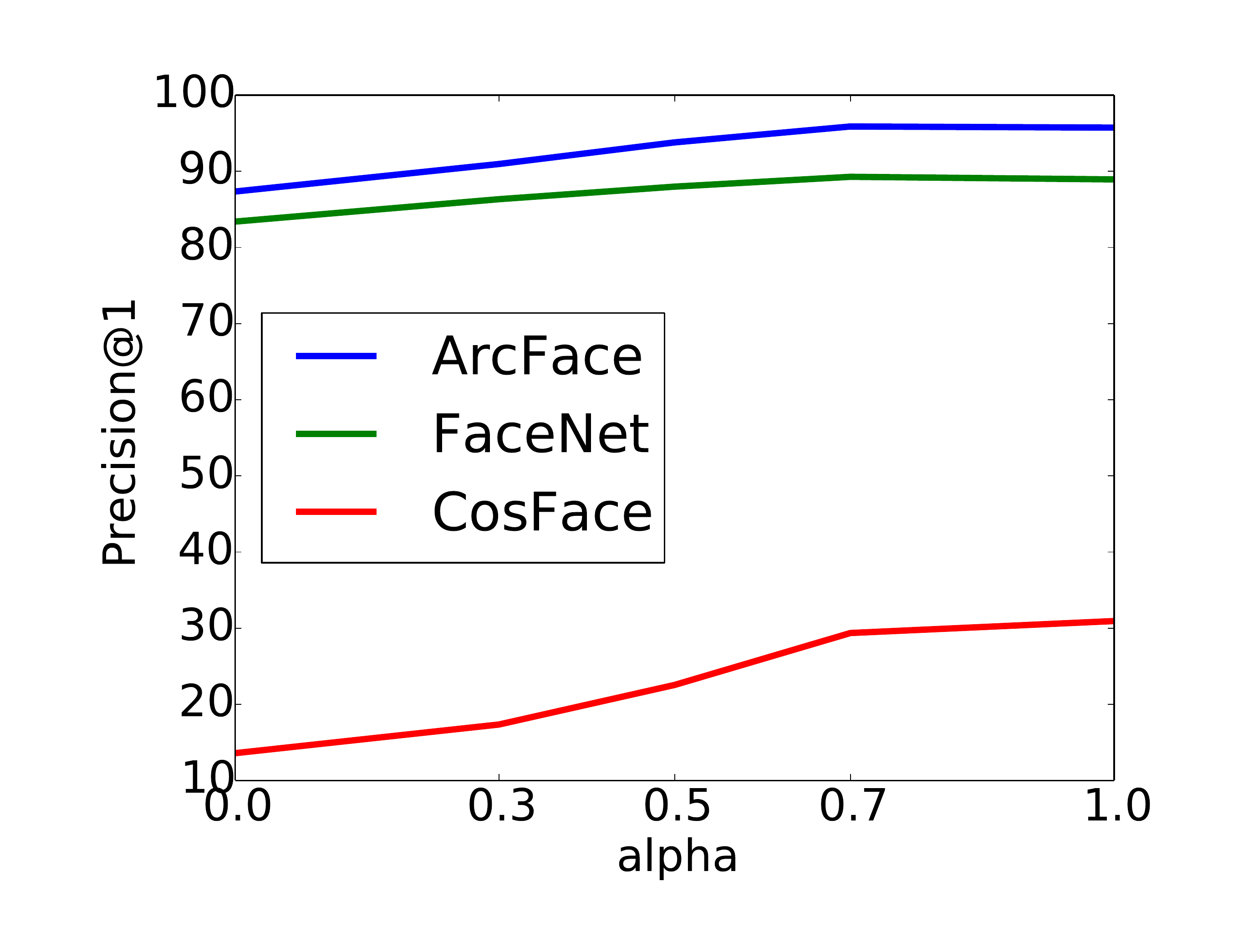}
        \caption{APC (LFW-crop)}
        \label{fig:apc_crop}
    \end{subfigure}\bigskip 
    \begin{subfigure}{0.32\textwidth}
        \includegraphics[width=\linewidth]{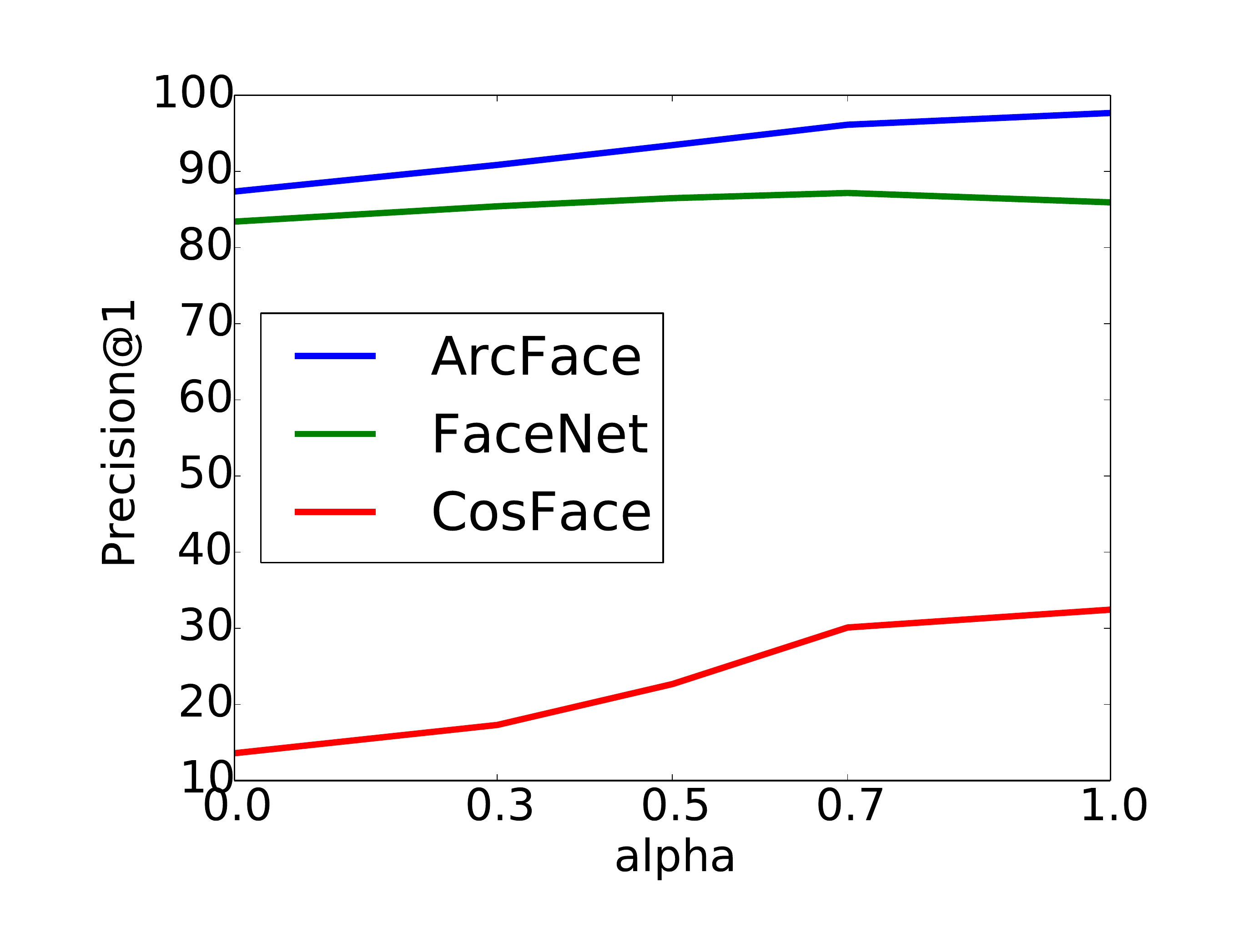}
        \caption{Uniform (LFW-crop)}
        \label{fig:uniform_crop}
    \end{subfigure}\bigskip 
    \begin{subfigure}{0.32\textwidth}
        \includegraphics[width=\linewidth]{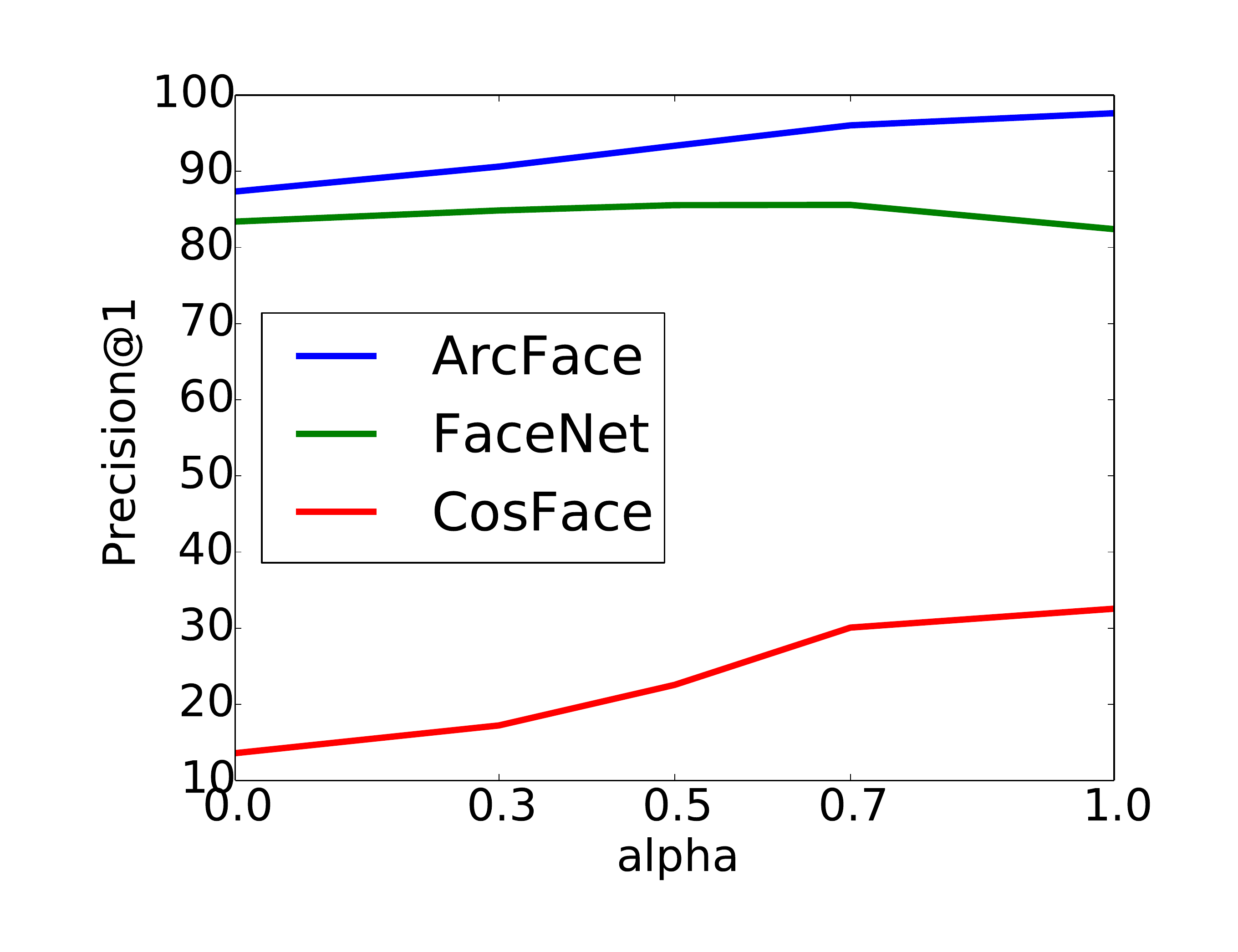}
        \caption{SC (LFW-crop)}
        \label{fig:sc_crop}
    \end{subfigure}
    \vspace*{-10mm}
\caption{
The P@1 of our 2-stage FI when sweeping across $\alpha \in \{ 0, 0.3, 0.5, 0.7, 1.0 \}$ for linearly combining EMD and cosine distance on LFW (top row; a--c) and LFW-crop images (bottom row; d--f) of all feature weighting (APC, Uniform, and SC). 
}
\label{fig:supp_alpha_effect}
\end{figure*}

\newpage
\begin{table}[t]
\centering
\small
\hskip -0.3cm
\begin{tabular}{l|l|l|l|l|l}
\hline
Dataset                                          & Model                    & Method  & P@1    & RP    & M@R   \\ \hline
\multicolumn{1}{c|}{\multirow{6}{*}{\thead{CALFW \\ (Mask)}  }} & \multirow{4}{*}{\thead{ArcFace}} & Stage 1  & 96.81 & 53.13 & 51.70 \\ \cline{3-6} 
\multicolumn{1}{c|}{}                           &                          & APC     & \textbf{99.92} & \textbf{57.27} & \textbf{56.33} \\ 
\multicolumn{1}{c|}{}                           &                          & Uniform     & \textbf{99.92} & \textbf{57.28} & \textbf{56.24} \\ 
\multicolumn{1}{c|}{}                           &                          & SC     & \textbf{99.92} & \textbf{57.13} & \textbf{56.06} \\ \cline{2-6} 
\multicolumn{1}{c|}{}                           & \multirow{4}{*}{CosFace} & Stage 1  & 98.54 & 43.46 & 41.20 \\ \cline{3-6} 
\multicolumn{1}{c|}{}                           &                          & SC      & \textbf{99.96} & \textbf{59.87} & \textbf{58.93} \\ 
\multicolumn{1}{c|}{}                           &                          & Uniform      & \textbf{99.96} & \textbf{59.86} & \textbf{58.91} \\ 
\multicolumn{1}{c|}{}                           &                          & APC      & \textbf{99.96} & \textbf{59.85} & \textbf{58.87} \\ \cline{2-6} 
\multicolumn{1}{c|}{}                           & \multirow{4}{*}{FaceNet} & Stage 1  & 77.63 & 39.74 & 36.93 \\ \cline{3-6} 
\multicolumn{1}{c|}{}                           &                          & APC     & \textbf{96.67} & \textbf{45.87} & \textbf{44.53} \\ 
\multicolumn{1}{c|}{}                           &                          & Uniform     & \textbf{94.23} & \textbf{43.90} & \textbf{42.33} \\ 
\multicolumn{1}{c|}{}                           &                          & SC     & \textbf{90.80} & \textbf{42.85} & \textbf{40.95} \\ \hline
\multirow{6}{*}{\thead{CALFW \\ (Sunglass)}}                       & \multirow{4}{*}{ArcFace} & Stage 1  & 51.11 & 29.38 & 26.73 \\ \cline{3-6} 
                                                 &                          & Uniform & \textbf{55.80} & \textbf{31.50} & \textbf{28.60} \\ 
                                                 &                          & APC & \textbf{54.95} & \textbf{30.66} & \textbf{27.74} \\ 
                                                 &                          & SC & \textbf{55.45} & \textbf{31.42} & \textbf{28.49} \\ \cline{2-6} 
                                                 & \multirow{4}{*}{CosFace} & Stage 1  & 45.20 & 25.93 & 22.78 \\ \cline{3-6} 
                                                 &                          & Uniform & \textbf{50.28} & \textbf{27.23} & \textbf{24.40} \\ 
                                                 &                          & APC & \textbf{49.67} & \textbf{26.98} & \textbf{24.12} \\ 
                                                 &                          & SC & \textbf{50.24} & \textbf{27.22} & \textbf{24.38} \\ \cline{2-6} 
                                                 & \multirow{4}{*}{FaceNet} & Stage 1  & 21.68 & 13.70 & 10.89 \\ \cline{3-6} 
                                                 &                          & APC     & \textbf{25.07} & \textbf{15.04} & \textbf{12.16} \\ 
                                                 &                          & Uniform     & \textbf{25.08} & \textbf{14.97} & \textbf{12.21} \\ 
                                                 &                          & SC     & \textbf{24.38} & \textbf{14.58} & \textbf{11.88} \\ \hline
\multirow{6}{*}{\thead{CALFW \\ (Crop)}}                       & \multirow{4}{*}{ArcFace} & Stage 1  & 79.13 & 43.46 & 41.20 \\ \cline{3-6} 
                                                 &                          & Uniform & \textbf{94.04} & \textbf{49.57} & \textbf{48.15} \\ 
                                                 &                          & APC & \textbf{92.57} & \textbf{47.17} & \textbf{45.68} \\ 
                                                 &                          & SC & \textbf{93.76} & \textbf{49.51} & \textbf{48.05} \\ \cline{2-6} 
                                                 & \multirow{4}{*}{CosFace} & Stage 1  & 10.99 & 6.45  & 5.43  \\ \cline{3-6} 
                                                 &                          & SC      & \textbf{27.42} & \textbf{12.68} & \textbf{11.59} \\ 
                                                 &                          & Uniform      & \textbf{27.43} & \textbf{12.66} & \textbf{11.58} \\ 
                                                 &                          & APC      & \textbf{25.99} & \textbf{12.35} & \textbf{11.13} \\ \cline{2-6} 
                                                 & \multirow{4}{*}{FaceNet} & Stage 1  & 79.47 & 44.40 & 41.99 \\ \cline{3-6} 
                                                 &                          & APC     & \textbf{85.71} & \textbf{45.91} & \textbf{43.83} \\ 
                                                 &                          & Uniform     & \textbf{83.92} & \textbf{45.22} & \textbf{43.04} \\ 
                                                 &                          & SC     & \textbf{82.33} & \textbf{44.54} & \textbf{42.26} \\ \hline
\end{tabular}
\caption{
Our 2-stage method for all feature weighting methods (APC, SC, and Uniform) for face occlusions (\eg mask, sunglass, and crop)  is substantially more robust to the Stage 1 alone baseline (ST1) on CALFW \cite{Tianyue2017calfw}.
}

\label{tab:face_occ_calfw_full_re}
\end{table}

\begin{table}[t]
\centering
\small
\hskip -0.3cm
\begin{tabular}{l|l|l|l|l|l}
\hline
Dataset                                          & Model                    & Method  & P@1    & RP    & M@R   \\ \hline
\multicolumn{1}{c|}{\multirow{6}{*}{\thead{AgeDB \\ (Mask)}}} & \multirow{4}{*}{ArcFace} & Stage 1  & 96.15 & 39.22 & 30.41 \\ \cline{3-6} 
                                                &                          & APC     & \textbf{99.84} & \textbf{39.22} & \textbf{33.18} \\ 
                                                &                          & Uniform     & \textbf{99.82} & \textbf{39.23} & \textbf{32.94} \\ 
                                                &                          & SC     & \textbf{99.82} & \textbf{39.12} & \textbf{32.77} \\ \cline{2-6} 
                                                & \multirow{4}{*}{CosFace} & Stage 1  & 98.31 & 38.17 & 31.57 \\ \cline{3-6} 
                                                &                          & APC     & \textbf{99.95} & \textbf{39.70} & \textbf{33.68} \\ 
                                                &                          & Uniform     & \textbf{99.95} & \textbf{39.61} & \textbf{33.60} \\ 
                                                &                          & SC     & \textbf{99.95} & \textbf{39.63} & \textbf{33.62} \\ \cline{2-6} 
                                                & \multirow{4}{*}{FaceNet} & Stage 1  & 75.99 & 22.28 & 14.95 \\ \cline{3-6} 
                                                &                          & APC     & \textbf{96.53} & \textbf{24.25} & \textbf{17.49} \\ 
                                                &                          & Uniform     & \textbf{93.99} & \textbf{22.55} & \textbf{15.68} \\ 
                                                &                          & SC     & \textbf{90.60} & \textbf{22.14} & \textbf{15.13} \\ \hline
\multicolumn{1}{c|}{\multirow{6}{*}{\thead{AgeDB \\ (Sunglass)}}} & \multirow{4}{*}{ArcFace} & Stage 1  & 84.64 & 51.16 & 44.99 \\ \cline{3-6} 
                                                                  &                          & Uniform     & \textbf{88.06} & \textbf{51.17} & \textbf{45.24} \\ 
                                                                  &                          & APC     & \textbf{87.06} & \textbf{50.40} & \textbf{44.27} \\ 
                                                                  &                          & SC     & \textbf{87.96} & \textbf{51.16} & \textbf{45.22} \\ \cline{2-6} 
                                                                  & \multirow{4}{*}{CosFace} & Stage 1  & 68.93 & 34.90 & 27.30 \\ \cline{3-6} 
                                                                  &                          & APC     & \textbf{75.97} & \textbf{35.54} & \textbf{28.12} \\ 
                                                                  &                          & Uniform     & \textbf{74.85} & \textbf{35.33} & \textbf{27.79} \\ 
                                                                  &                          & SC     & \textbf{74.82} & \textbf{35.33} & \textbf{27.79} \\ \cline{2-6} 
                                    & \multirow{4}{*}{FaceNet} & Stage 1  & 56.77 & 27.92 & 20.00 \\ \cline{3-6} 
                                    &                          & APC     & \textbf{61.21} & \textbf{28.98} & \textbf{21.11} \\ 
                                    &                          & Uniform     & \textbf{61.64} & \textbf{28.62} & \textbf{20.94} \\ 
                                    &                          & SC     & \textbf{61.27} & \textbf{28.44} & \textbf{20.76} \\ \hline
\multicolumn{1}{c|}{\multirow{6}{*}{\thead{AgeDB \\ (Crop)}}} & \multirow{4}{*}{ArcFace} & Stage 1  & 79.92 & 32.66 & 26.19 \\ \cline{3-6} 
                                                              &                          & Uniform     & \textbf{94.18} & \textbf{34.81} & \textbf{28.80} \\ 
                                                              &                          & APC     & \textbf{92.92} & \textbf{32.93} & \textbf{26.60} \\ 
                                                              &                          & SC     & \textbf{94.03} & \textbf{34.83} & \textbf{28.80} \\ \cline{2-6} 
                     & \multirow{4}{*}{CosFace} & Stage 1  & 10.11 & 4.23 & 2.18 \\ \cline{3-6} 
                     &                          & SC     & \textbf{21.00} & \textbf{5.02} & \textbf{2.89} \\ 
                     &                          & Uniform     & \textbf{20.96} & \textbf{5.02} & \textbf{2.88} \\ 
                     &                          & APC     & \textbf{19.58} & \textbf{4.95} & \textbf{2.76} \\ \cline{2-6} 
                     & \multirow{2}{*}{FaceNet} & Stage 1  & 80.80 & 31.50 & 24.27 \\ \cline{3-6} 
                     &                          & APC     & \textbf{86.74} & \textbf{31.51} & \textbf{24.32} \\ 
                     &                          & Uniform     & \textbf{84.93} & \textbf{30.87} & \textbf{23.68} \\ 
                     &                          & SC     & \textbf{83.29} & \textbf{30.51} & \textbf{23.24} \\ \hline
\end{tabular}
\caption{
Our 2-stage method for all feature weighting methods (APC, SC, and Uniform) for face occlusions (\eg mask, sunglass, and crop) is substantially more robust to the Stage 1 alone baseline (ST1) on AgeDB \cite{moschoglou2017agedb}.
}
\label{tab:face_occ_agedb_full_re}
\end{table}

\begin{table}[t]
\centering
\small
\hskip -0.3cm
\begin{tabular}{l|l|l|l|l|l}
\hline
Dataset                                          & Model                    & Method  & P@1    & RP    & M@R   \\ \hline
\multicolumn{1}{c|}{\multirow{12}{*}{\thead{CFP \\ (Mask)}}} & \multirow{4}{*}{ArcFace} & Stage 1  & 96.65 & 69.88 & 66.67 \\ \cline{3-6} 
                                                &                          & APC     & \textbf{99.78} & \textbf{76.07} & \textbf{74.20} \\ 
                                                &                          & Uniform     & \textbf{99.78} & \textbf{76.41} & \textbf{74.34} \\ 
                                                &                          & SC     & \textbf{99.78} & \textbf{76.23} & \textbf{74.08} \\ \cline{2-6} 
                                                & \multirow{4}{*}{CosFace} & Stage 1  & 92.52 & 66.14 & 62.73 \\ \cline{3-6} 
                                                &                          & APC     & \textbf{94.22} & \textbf{69.56} & \textbf{66.66} \\ 
                                                &                          & Uniform     & \textbf{94.38} & \textbf{70.34} & \textbf{67.59} \\ 
                                                &                          & SC     & \textbf{94.32} & \textbf{70.45} & \textbf{67.72} \\ \cline{2-6} 
                                                & \multirow{4}{*}{FaceNet} & Stage 1  & 83.96 & 54.82 & 49.01 \\ \cline{3-6} 
                                                &                          & APC     & \textbf{97.48} & \textbf{61.58} & \textbf{57.35} \\ 
                                                &                          & Uniform     & \textbf{95.63} & \textbf{58.71} & \textbf{53.96} \\ 
                                                &                          & SC     & \textbf{93.09} & \textbf{57.30} & \textbf{52.15} \\ \hline
\multicolumn{1}{c|}{\multirow{12}{*}{\thead{CFP \\ (Sunglass)}}} & \multirow{4}{*}{ArcFace} & Stage 1  & 91.54 & 70.63 & 67.21 \\ \cline{3-6} 
                                                                  &                          & Uniform     & \textbf{93.10} & \textbf{71.75} & \textbf{68.33} \\ 
                                                                  &                          & APC     & \textbf{94.06} & \textbf{71.05} & \textbf{67.89} \\ 
                                                                  &                          & SC     & \textbf{92.92} & \textbf{71.69} & \textbf{68.24} \\ \cline{2-6} 
                                                                  & \multirow{4}{*}{CosFace} & Stage 1  & 88.72 & 65.93 & 61.97 \\ \cline{3-6} 
                                                                  &                          & APC     & \textbf{82.22} & \textbf{60.33} & \textbf{54.25} \\ 
                                                                  &                          & Uniform     & \textbf{85.28} & \textbf{61.89} & \textbf{56.65} \\ 
                                                                  &                          & SC     & \textbf{86.04} & \textbf{62.53} & \textbf{57.45} \\ \cline{2-6} 
                                    & \multirow{4}{*}{FaceNet} & Stage 1  & 69.02 & 50.58 & 43.26 \\ \cline{3-6} 
                                    &                          & APC     & \textbf{74.98} & \textbf{52.98} & \textbf{46.14} \\ 
                                    &                          & Uniform     & \textbf{69.18} & \textbf{51.46} & \textbf{43.87} \\ 
                                    &                          & SC     & \textbf{67.90} & \textbf{50.67} & \textbf{43.02} \\ \hline
\multicolumn{1}{c|}{\multirow{12}{*}{\thead{CFP \\ (Crop)}}} & \multirow{4}{*}{ArcFace} & Stage 1  & 91.34 & 65.13 & 61.37 \\ \cline{3-6} 
                                                              &                          & Uniform     & \textbf{98.16} & \textbf{70.77} & \textbf{67.80} \\ 
                                                              &                          & APC     & \textbf{97.96} & \textbf{67.51} & \textbf{64.15} \\ 
                                                              &                          & SC     & \textbf{98.04} & \textbf{70.78} & \textbf{67.78} \\ \cline{2-6} 
                     & \multirow{4}{*}{CosFace} & Stage 1  & 17.06 & 10.51 & 8.02 \\ \cline{3-6} 
                     &                          & SC     & \textbf{34.60} & \textbf{15.69} & \textbf{12.96} \\ 
                     &                          & Uniform     & \textbf{34.50} & \textbf{15.63} & \textbf{12.90} \\ 
                     &                          & APC     & \textbf{32.22} & \textbf{15.07} & \textbf{12.23} \\ \cline{2-6} 
                     & \multirow{4}{*}{FaceNet} & Stage 1  & 95.20 & 72.70 & 69.43 \\ \cline{3-6} 
                     &                          & APC     & \textbf{97.34} & \textbf{72.63} & \textbf{69.47} \\ 
                     &                          & Uniform     & \textbf{96.54} & \textbf{72.78} & \textbf{69.56} \\ 
                     &                          & SC     & \textbf{96.02} & \textbf{72.22} & \textbf{68.88} \\ \hline
\multicolumn{1}{c|}{\multirow{12}{*}{\thead{CFP \\ (Profile)}}} & \multirow{4}{*}{ArcFace} & Stage 1  & 84.84 & 71.09 & 67.35 \\ \cline{3-6} 
                                                              &                          & Uniform     & \textbf{86.13} & \textbf{72.19} & \textbf{68.58} \\ 
                                                              &                          & APC     & \textbf{85.56} & \textbf{71.60} & \textbf{67.84} \\ 
                                                              &                          & SC     & \textbf{86.18} & \textbf{72.22} & \textbf{68.59} \\ \cline{2-6} 
                     & \multirow{4}{*}{CosFace} & Stage 1  & 71.64 & 58.87 & 54.81 \\ \cline{3-6} 
                     &                          & SC     & \textbf{71.74} & \textbf{59.27} & \textbf{55.27} \\ 
                     &                          & Uniform     & \textbf{71.74} & \textbf{59.21} & \textbf{55.22} \\ 
                     &                          & APC     & \textbf{71.64} & \textbf{59.24} & \textbf{55.23} \\ \cline{2-6} 
                     & \multirow{4}{*}{FaceNet} & Stage 1  & 75.71 & 61.78 & 56.30 \\ \cline{3-6} 
                     &                          & APC     & \textbf{76.38} & 61.69 & 56.19 \\ 
                     &                          & Uniform     & \textbf{76.33} & 61.47 & 55.89 \\ 
                     &                          & SC     & \textbf{76.22} & 61.35 & 55.74 \\ \hline
\end{tabular}
\caption{
More results of our 2-stage approach based on ArcFace features (8$\times$8 grid), CosFace features (6$\times$ 7), and FaceNet features (3 $\times$ 3) across all feature weighting methods which perform slightly better than the Stage 1 alone (ST1) baseline at P@1 when the query is a rotated face (\ie profile faces from CFP \cite{Sengupta2016cfp}).
}
\label{tab:face_occ_cfp_full_re}
\end{table}

\begin{table}[t]
\centering
\small
\hskip -0.3cm
\begin{tabular}{l|l|l|l|l|l}
\hline
Dataset                                          & Model                    & Method  & P@1    & RP    & M@R   \\ \hline
\multicolumn{1}{c|}{\multirow{12}{*}{\thead{TALFW}}} & \multirow{4}{*}{ArcFace} & Cosine  & 93.49 & 81.04 & 80.35 \\ \cline{3-6} 
                                                              &                          & Uniform     & \textbf{96.72} & \textbf{83.41} & \textbf{82.80} \\ 
                                                              &                          & APC     & \textbf{96.54} & \textbf{82.72} & \textbf{82.10} \\ 
                                                              &                          & SC     & \textbf{96.71} & \textbf{83.39} & \textbf{82.78} \\ \cline{2-6} 
                     & \multirow{4}{*}{CosFace} & Cosine  & 96.49 & 83.57 & 82.99 \\ \cline{3-6} 
                     &                          & SC     & \textbf{99.14} & \textbf{85.03} & \textbf{55.27} \\ 
                     &                          & Uniform     & \textbf{99.14} & \textbf{85.56} & \textbf{85.11} \\ 
                     &                          & APC     & \textbf{99.07} & \textbf{85.48} & \textbf{85.08} \\ \cline{2-6} 
                     & \multirow{4}{*}{FaceNet} & Cosine  & 95.33 & 79.24 & 78.19 \\ \cline{3-6} 
                     &                          & APC     & \textbf{97.26} & \textbf{80.33} & \textbf{79.39} \\ 
                     &                          & Uniform     & \textbf{97.70} & \textbf{80.10} & \textbf{79.15} \\ 
                     &                          & SC     & \textbf{97.59} & \textbf{79.85} & \textbf{78.89} \\ \hline
\end{tabular}
\caption{
Our re-ranking consistently improves the precision over Stage 1 alone (ST1) when identifying adversarial TALFW \cite{zhong2020towards} images given an in-distribution LFW \cite{yi2014learning} gallery. The conclusions also carry over to other feature-weighting methods and models (ArcFace, CosFace, FaceNet).
}
\label{tab:face_occ_talfw_full_re}
\end{table}



\section{Additional Results: Face Verification on MLFW}
\label{sec:verification}
In the main text, we find that DeepFace-EMD is effective in face \emph{identification} given many types of OOD images.
Here, we also evaluate DeepFace-EMD for face \emph{verification} of MLFW \cite{wang2021mlfw}, a recent benchmark that consists of masked LFW faces. 
As in common verification setups of LFW \cite{schroff2015facenet, Liu_2017_CVPR,wang2021mlfw}, given pairs of face images and their similarity scores predicted by a verification system, we find the optimal threshold that yields the best accuracy.
Here, we follow the setup in \cite{wang2021mlfw} to enable a fair comparison. 
First of all,  we reproduce Table 3 in \cite{wang2021mlfw}, which evaluate face verification accuracy on 6,000 pair of MLFW images. 
Then, we run our DeepFace-EMD distance function (Eq.~\ref{eq:alpha}). We found that using our proposed distance consistently improves on face \emph{verification} for all three PyTorch models in \cite{wang2021mlfw}.
Interestingly, with DeepFace-EMD, \textbf{we obtained a state-of-the-art result} (91.17\%) on MLFW (see \cref{tab:mlfw_fr_comparison}).

\begin{table}[h]
\small
\centering
\begin{tabular}{l|l|l}
\hline
Models in MLFW Table 3 [58]             & Method  & MLFW     \\ \hline
\multirow{2}{*}{\thead{Private-Asia, R50, ArcFace}} & [58]  & 74.85\% \\ 
                         & {+ DeepFaceEMD}     & \textbf{76.50\%} \\ \cline{1-3} 
\multirow{2}{*}{\thead{CASIA, R50, CosFace}} & [58]  & 82.87\% \\ 
                          & {+ DeepFaceEMD}     & \textbf{87.17\%} \\ \cline{1-3} 
\multirow{2}{*}{\thead{MS1MV2, R100, Curricularface}} & [58]  & 90.60\% \\ 
                          & {+ DeepFaceEMD}     & \textbf{91.17\%} \\ \hline
\end{tabular}
\caption{
Using our proposed similarity function consistently improves the face verification results on MLFW (\ie OOD masked images) for models reported in Wang et al.~\cite{wang2021mlfw}.
We use pre-trained models and code by \cite{wang2021mlfw}. 
}
\label{tab:mlfw_fr_comparison}
\end{table}

\section{Additional ablation studies: 3D Facial Alignment vs. MTCNN}
\label{sec:mtcnn}

The reason we used the 3D alignment pre-processing instead of the default MTCNN pre-processing \cite{Zhang2016IEEE} of the three models was because for ArcFace, the 3D alignment actually resulted in better P@1, RP, and M@R for both our baselines and DeepFace-EMD (\eg +3.35\% on MLFW).
For FaceNet, the 3D alignment did yield worse performance compared to MTCNN.
However, we confirm that our conclusions that \textbf{DeepFace-EMD improves FI on the reported datasets regardless of the pre-processing choice}. 
See \cref{tab:compare_mtcnn}  for details.
\begin{table}[t]
\small
\hskip 0.25cm
\centering
\begin{tabular}{l|l|l|l|l|l|l}
\hline
Dataset                                          & Model & Pre-processing                    & Method  & P@1    & RP    & M@R   \\ \hline
\multicolumn{1}{c|}{\multirow{8}{*}{\thead{CALFW \\ (Mask)}  }} & \multirow{4}{*}{\thead{ArcFace}} & \multirow{2}{*}{\thead{3D alignment}}  & ST1   & 96.81 & 53.13 & 51.70 \\ 
\multicolumn{1}{c|}{}                           &                          &  &Ours     & \textbf{99.92} & \textbf{57.27} & \textbf{56.33} \\ \cline{3-7} 
\multicolumn{1}{c|}{}                           &  & \multirow{2}{*}{\thead{MTCNN}} &ST1  & 96.36 & 48.35 & 46.85 \\ 
\multicolumn{1}{c|}{}                           &                          &  &Ours      & \textbf{99.92} & \textbf{53.53} & \textbf{52.53} \\ \cline{2-7} 
\multicolumn{1}{c|}{}                           & \multirow{4}{*}{FaceNet} & \multirow{2}{*}{\thead{3D alignment}} &ST1   & 77.63 & 39.74 & 36.93 \\ 
\multicolumn{1}{c|}{}                           &                          &  & Ours     & \textbf{96.67} & \textbf{45.87} & \textbf{44.53} \\ \cline{3-7}
\multicolumn{1}{c|}{}                           &  & \multirow{2}{*}{\thead{MTCNN}} &ST1   & 86.65 & 45.29 & 42.83 \\ 
\multicolumn{1}{c|}{}                           &                          &  & Ours     & \textbf{98.62} & \textbf{49.75} & \textbf{48.49} \\ \hline
\multicolumn{1}{c|}{\multirow{8}{*}{\thead{AgeDB \\ (Mask)}}} & \multirow{4}{*}{ArcFace} & \multirow{2}{*}{\thead{3D alignment}} & ST1   & 96.15 & 39.22 & 30.41 \\ 
\multicolumn{1}{c|}{}                           &                         &  & Ours     & \textbf{99.84} & {39.22} & \textbf{33.18} \\ \cline{3-7} 
\multicolumn{1}{c|}{}                           & & \multirow{2}{*}{\thead{MTCNN}} & ST1   & 95.35 & 29.51 & 22.75 \\ 
\multicolumn{1}{c|}{}                           &                          &  & Ours     & \textbf{99.78} & \textbf{32.82} & \textbf{27.08} \\ \cline{2-7} 
\multicolumn{1}{c|}{}                           & \multirow{4}{*}{FaceNet} & \multirow{2}{*}{\thead{3D alignment}} & ST1   & 75.99 & 22.28 & 14.95 \\ 
\multicolumn{1}{c|}{}                           &                         &  & Ours     & \textbf{96.53} & \textbf{24.25} & \textbf{17.49} \\ \cline{3-7}
\multicolumn{1}{c|}{}                           &  & \multirow{2}{*}{\thead{MTCNN}} & ST1   & 83.93 & 25.18 & 17.74 \\ 
\multicolumn{1}{c|}{}                           &                         &  & Ours     & \textbf{98.26} & \textbf{27.27} & \textbf{20.45} \\ \hline
\end{tabular}
\caption{
DeepFace-EMD improved FI on the reported datasets regardless of the pre-processing choice.
}
\label{tab:compare_mtcnn}
\end{table}

\begin{table}[t]
\small
\centering
\resizebox{7.3cm}{!}{%
\begin{tabular}{l|l|l|l|l|l}
\hline
Dataset                                          & Model                    & Method  & P@1    & RP    & M@R   \\ \hline
\multicolumn{1}{c|}{\multirow{6}{*}{\thead{CFP \\ (Profile)}}} & \multirow{2}{*}{ArcFace} & ST1   & 84.84 & 71.09 & 67.35 \\ 
           &                          & Ours     & \textbf{84.94} & {70.31} & {66.36} \\ \cline{2-6} 
\multicolumn{1}{c|}{}                           & \multirow{2}{*}{CosFace} & ST1   & 71.64 & 58.87 & 54.81 \\ 
&                          & Ours     & {71.64} & \textbf{59.24} & \textbf{55.23} \\ \cline{2-6} 
\multicolumn{1}{c|}{}                           & \multirow{2}{*}{FaceNet} & ST1   & 75.71 & 61.78 & 56.30 \\ 
\multicolumn{1}{c|}{}                           &                          & Ours     & \textbf{76.38} & {61.69} & {56.19} \\ \hline
\end{tabular}
}
\caption{
Our 2-stage approach based on ArcFace features (8$\times$8 grid; APC) performs slightly better than the Stage 1 alone (ST1) baseline at P@1 when the query is a rotated face (\ie profile faces from CFP \cite{Sengupta2016cfp}).
See \cref{tab:face_occ_cfp_full_re} for the results of occlusions on CFP.
}
\label{tab:face_cfp_occ_re}
\end{table}

\end{document}

%% file: math_commands.tex
\usepackage{amsmath,amsfonts,bm}

\def\1{\bm{1}}

\def\mD{{\bm{D}}}

\def\mF{{\bm{F}}}

\DeclareMathAlphabet{\mathsfit}{\encodingdefault}{\sfdefault}{m}{sl}
\SetMathAlphabet{\mathsfit}{bold}{\encodingdefault}{\sfdefault}{bx}{n}

\def\gG{{\mathcal{G}}}

\def\gQ{{\mathcal{Q}}}

\def\sR{{\mathbb{R}}}












%% file: main.bbl
\begin{thebibliography}{10}\itemsep=-1pt

\bibitem{airplane}
Facial recognition tech at hartsfield-jackson wins over most international
  delta customers - atlanta business chronicle.
\newblock
  \url{https://www.bizjournals.com/atlanta/news/2019/06/25/facial-recognition-tech-at-hartsfield-jackson-wins.html}.
\newblock (Accessed on 11/09/2021).

\bibitem{newjersey2021arrested}
Flawed facial recognition leads to arrest and jail for new jersey man - the new
  york times.
\newblock
  \url{https://www.nytimes.com/2020/12/29/technology/facial-recognition-misidentify-jail.html}.
\newblock (Accessed on 11/09/2021).

\bibitem{michigan2021arrested}
Michigan man wrongfully accused with facial recognition urges congress to act.
\newblock
  \url{https://www.detroitnews.com/story/news/politics/2021/07/13/house-panel-hear-michigan-man-wrongfully-accused-facial-recognition/7948908002/}.
\newblock (Accessed on 11/09/2021).

\bibitem{lawsuit2021facial}
The new lawsuit that shows facial recognition is officially a civil rights
  issue | mit technology review.
\newblock
  \url{https://www.technologyreview.com/2021/04/14/1022676/robert-williams-facial-recognition-lawsuit-aclu-detroit-police/}.
\newblock (Accessed on 04/15/2021).

\bibitem{MiaSato.2021}
The pandemic is testing the limits of face recognition | mit technology review.
\newblock
  \url{https://www.technologyreview.com/2021/09/28/1036279/pandemic-unemployment-government-face-recognition/}.
\newblock (Accessed on 11/09/2021).

\bibitem{arcfacePyTorch}
ronghuaiyang/arcface-pytorch.
\newblock \url{https://github.com/ronghuaiyang/arcface-pytorch}.
\newblock (Accessed on 11/16/2021).

\bibitem{detroit2021arrested}
Wrongfully arrested man sues detroit police following false facial-recognition
  match - the washington post.
\newblock
  \url{https://www.washingtonpost.com/technology/2021/04/13/facial-recognition-false-arrest-lawsuit/}.
\newblock (Accessed on 11/09/2021).

\bibitem{amos2016openface}
Brandon Amos, Bartosz Ludwiczuk, and Mahadev Satyanarayanan.
\newblock Openface: A general-purpose face recognition library with mobile
  applications.
\newblock Technical report, CMU-CS-16-118, CMU School of Computer Science,
  2016.

\bibitem{anonymous2022patches}
Anonymous.
\newblock Patches are all you need?
\newblock In {\em Submitted to The Tenth International Conference on Learning
  Representations}, 2022.
\newblock under review.

\bibitem{Anwar2020MaskedFR}
Aqeel Anwar and Arijit Raychowdhury.
\newblock Masked face recognition for secure authentication.
\newblock {\em ArXiv}, abs/2008.11104, 2020.

\bibitem{Bhagavatula2017FasterTR}
Chandrasekhar Bhagavatula, Chenchen Zhu, Khoa Luu, and Marios Savvides.
\newblock Faster than real-time facial alignment: A 3d spatial transformer
  network approach in unconstrained poses.
\newblock {\em 2017 IEEE International Conference on Computer Vision (ICCV)},
  pages 4000--4009, 2017.

\bibitem{bharadwaj2014aiding}
Samarth Bharadwaj, Mayank Vatsa, and Richa Singh.
\newblock Aiding face recognition with social context association rule based
  re-ranking.
\newblock In {\em IEEE International Joint Conference on Biometrics}, pages
  1--8. IEEE, 2014.

\bibitem{cai2020semi}
Jiancheng Cai, Hu Han, Jiyun Cui, Jie Chen, Li Liu, and S~Kevin Zhou.
\newblock Semi-supervised natural face de-occlusion.
\newblock {\em IEEE Transactions on Information Forensics and Security},
  16:1044--1057, 2020.

\bibitem{cao2018vggface2}
Qiong Cao, Li Shen, Weidi Xie, Omkar~M. Parkhi, and Andrew Zisserman.
\newblock Vggface2: A dataset for recognising faces across pose and age.
\newblock In {\em 2018 13th IEEE International Conference on Automatic Face
  Gesture Recognition (FG 2018)}, pages 67--74, 2018.

\bibitem{Cao2018VGGFace2AD}
Qiong Cao, Li Shen, Weidi Xie, Omkar~M. Parkhi, and Andrew Zisserman.
\newblock Vggface2: A dataset for recognising faces across pose and age.
\newblock {\em 2018 13th IEEE International Conference on Automatic Face \&
  Gesture Recognition (FG 2018)}, pages 67--74, 2018.

\bibitem{cohen1999finding}
Scott Cohen.
\newblock {\em Finding color and shape patterns in images}.
\newblock stanford university, 1999.

\bibitem{cui2008real}
Jingyu Cui, Fang Wen, and Xiaoou Tang.
\newblock Real time google and live image search re-ranking.
\newblock In {\em Proceedings of the 16th ACM international conference on
  Multimedia}, pages 729--732, 2008.

\bibitem{NIPS2013_af21d0c9}
Marco Cuturi.
\newblock Sinkhorn distances: Lightspeed computation of optimal transport.
\newblock In C.~J.~C. Burges, L. Bottou, M. Welling, Z. Ghahramani, and K.~Q.
  Weinberger, editors, {\em Advances in Neural Information Processing Systems}.
  Curran Associates, Inc., 2013.

\bibitem{deng2018arcface}
Jiankang Deng, Jia Guo, Xue Niannan, and Stefanos Zafeiriou.
\newblock Arcface: Additive angular margin loss for deep face recognition.
\newblock In {\em CVPR}, 2019.

\bibitem{dong2020occlusion}
Jiayuan Dong, Liyan Zhang, Hanwang Zhang, and Weichen Liu.
\newblock Occlusion-aware gan for face de-occlusion in the wild.
\newblock In {\em 2020 IEEE International Conference on Multimedia and Expo
  (ICME)}, pages 1--6. IEEE, 2020.

\bibitem{Dong2020OpensetFI}
Xingbo Dong, Soohyung Kim, Zhe Jin, Jung~Yeon Hwang, Sangrae Cho, and A.B.J.
  Teoh.
\newblock Open-set face identification with index-of-max hashing by learning.
\newblock {\em Pattern Recognition}, 103:107277, 2020.

\bibitem{ge2020occluded}
Shiming Ge, Chenyu Li, Shengwei Zhao, and Dan Zeng.
\newblock Occluded face recognition in the wild by identity-diversity
  inpainting.
\newblock {\em IEEE Transactions on Circuits and Systems for Video Technology},
  30(10):3387--3397, 2020.

\bibitem{guo2018face}
Jianzhu Guo, Xiangyu Zhu, Zhen Lei, and Stan~Z Li.
\newblock Face synthesis for eyeglass-robust face recognition.
\newblock In {\em Chinese Conference on biometric recognition}, pages 275--284.
  Springer, 2018.

\bibitem{he2016identity}
Kaiming He, Xiangyu Zhang, Shaoqing Ren, and Jian Sun.
\newblock Identity mappings in deep residual networks.
\newblock In {\em European conference on computer vision}, pages 630--645.
  Springer, 2016.

\bibitem{he2011regularized}
Ran He, Wei-Shi Zheng, Bao-Gang Hu, and Xiang-Wei Kong.
\newblock A regularized correntropy framework for robust pattern recognition.
\newblock {\em Neural computation}, 23(8):2074--2100, 2011.

\bibitem{dlib09}
Davis~E. King.
\newblock Dlib-ml: A machine learning toolkit.
\newblock {\em Journal of Machine Learning Research}, 10:1755--1758, 2009.

\bibitem{kumar2017earth}
Sachin Kumar, Soumen Chakrabarti, and Shourya Roy.
\newblock Earth mover's distance pooling over siamese lstms for automatic short
  answer grading.
\newblock In {\em IJCAI}, pages 2046--2052, 2017.

\bibitem{kurakin2016adversarial}
Alexey Kurakin, Ian Goodfellow, Samy Bengio, et~al.
\newblock Adversarial examples in the physical world, 2016.

\bibitem{kusner2015word}
Matt Kusner, Yu Sun, Nicholas Kolkin, and Kilian Weinberger.
\newblock From word embeddings to document distances.
\newblock In {\em International conference on machine learning}, pages
  957--966. PMLR, 2015.

\bibitem{levina2001earth}
Elizaveta Levina and Peter Bickel.
\newblock The earth mover's distance is the mallows distance: Some insights
  from statistics.
\newblock In {\em Proceedings Eighth IEEE International Conference on Computer
  Vision. ICCV 2001}, volume~2, pages 251--256. IEEE, 2001.

\bibitem{li2013structured}
Xiao-Xin Li, Dao-Qing Dai, Xiao-Fei Zhang, and Chuan-Xian Ren.
\newblock Structured sparse error coding for face recognition with occlusion.
\newblock {\em IEEE transactions on image processing}, 22(5):1889--1900, 2013.

\bibitem{li2005nonparametric}
Zhifeng Li, Wei Liu, Dahua Lin, and Xiaoou Tang.
\newblock Nonparametric subspace analysis for face recognition.
\newblock In {\em 2005 IEEE Computer Society Conference on Computer Vision and
  Pattern Recognition (CVPR'05)}, volume~2, pages 961--966. IEEE, 2005.

\bibitem{Liu_2017_CVPR}
Weiyang Liu, Yandong Wen, Zhiding Yu, Ming Li, Bhiksha Raj, and Le Song.
\newblock Sphereface: Deep hypersphere embedding for face recognition.
\newblock In {\em The IEEE Conference on Computer Vision and Pattern
  Recognition (CVPR)}, 2017.

\bibitem{lupu2017new}
Noam Lupu, Luc{\'\i}a Selios, and Zach Warner.
\newblock A new measure of congruence: The earth mover’s distance.
\newblock {\em Political Analysis}, 25(1):95--113, 2017.

\bibitem{mahmood2021robustness}
Kaleel Mahmood, Rigel Mahmood, and Marten Van~Dijk.
\newblock On the robustness of vision transformers to adversarial examples.
\newblock {\em arXiv preprint arXiv:2104.02610}, 2021.

\bibitem{min2011improving}
Rui Min, Abdenour Hadid, and Jean-Luc Dugelay.
\newblock Improving the recognition of faces occluded by facial accessories.
\newblock In {\em 2011 IEEE International Conference on Automatic Face \&
  Gesture Recognition (FG)}, pages 442--447. IEEE, 2011.

\bibitem{moschoglou2017agedb}
Stylianos Moschoglou, Athanasios Papaioannou, Christos Sagonas, Jiankang Deng,
  Irene Kotsia, and Stefanos Zafeiriou.
\newblock Agedb: the first manually collected, in-the-wild age database.
\newblock In {\em Proceedings of the IEEE Conference on Computer Vision and
  Pattern Recognition Workshop}, volume~2, page~5, 2017.

\bibitem{musgrave2020metric}
Kevin Musgrave, Serge Belongie, and Ser-Nam Lim.
\newblock A metric learning reality check.
\newblock In {\em ECCV}, pages 681--699. Springer, 2020.

\bibitem{nguyen2015deep}
Anh Nguyen, Jason Yosinski, and Jeff Clune.
\newblock Deep neural networks are easily fooled: High confidence predictions
  for unrecognizable images.
\newblock In {\em Proceedings of the IEEE conference on computer vision and
  pattern recognition}, pages 427--436, 2015.

\bibitem{oh2008occlusion}
Hyun~Jun Oh, Kyoung~Mu Lee, and Sang~Uk Lee.
\newblock Occlusion invariant face recognition using selective local
  non-negative matrix factorization basis images.
\newblock {\em Image and Vision computing}, 26(11):1515--1523, 2008.

\bibitem{osherov2017increasing}
Elad Osherov and Michael Lindenbaum.
\newblock Increasing cnn robustness to occlusions by reducing filter support.
\newblock In {\em Proceedings of the IEEE International Conference on Computer
  Vision}, pages 550--561, 2017.

\bibitem{Peng2019TIP}
Chunlei Peng, Nannan Wang, Jie Li, and Xinbo Gao.
\newblock Re-ranking high-dimensional deep local representation for nir-vis
  face recognition.
\newblock {\em IEEE Transactions on Image Processing}, 28(9):4553--4565, 2019.

\bibitem{peng2006using}
Yuxin Peng, Cuihua Fang, and Xiaoou Chen.
\newblock Using earth mover’s distance for audio clip retrieval.
\newblock In {\em Pacific-Rim Conference on Multimedia}, pages 405--413.
  Springer, 2006.

\bibitem{qiu2021end2end}
Haibo Qiu, Dihong Gong, Zhifeng Li, Wei Liu, and Dacheng Tao.
\newblock End2end occluded face recognition by masking corrupted features.
\newblock {\em IEEE Transactions on Pattern Analysis and Machine Intelligence},
  2021.

\bibitem{rubner2000earth}
Yossi Rubner, Carlo Tomasi, and Leonidas~J Guibas.
\newblock The earth mover's distance as a metric for image retrieval.
\newblock {\em International journal of computer vision}, 40(2):99--121, 2000.

\bibitem{sarfraz2018pose}
M~Saquib Sarfraz, Arne Schumann, Andreas Eberle, and Rainer Stiefelhagen.
\newblock A pose-sensitive embedding for person re-identification with expanded
  cross neighborhood re-ranking.
\newblock In {\em Proceedings of the IEEE Conference on Computer Vision and
  Pattern Recognition}, pages 420--429, 2018.

\bibitem{schroff2015facenet}
Florian Schroff, Dmitry Kalenichenko, and James Philbin.
\newblock Facenet: A unified embedding for face recognition and clustering.
\newblock In {\em Proceedings of the IEEE conference on computer vision and
  pattern recognition}, pages 815--823, 2015.

\bibitem{Sengupta2016cfp}
S. Sengupta, J.C. Cheng, C.D. Castillo, V.M. Patel, R. Chellappa, and D.W.
  Jacobs.
\newblock Frontal to profile face verification in the wild.
\newblock {\em IEEE Conference on Applications of Computer Vision}, February
  2016.

\bibitem{shao2021adversarial}
Rulin Shao, Zhouxing Shi, Jinfeng Yi, Pin-Yu Chen, and Cho-Jui Hsieh.
\newblock On the adversarial robustness of visual transformers.
\newblock {\em arXiv preprint arXiv:2103.15670}, 2021.

\bibitem{sharif2016accessorize}
Mahmood Sharif, Sruti Bhagavatula, Lujo Bauer, and Michael~K Reiter.
\newblock Accessorize to a crime: Real and stealthy attacks on state-of-the-art
  face recognition.
\newblock In {\em Proceedings of the 2016 acm sigsac conference on computer and
  communications security}, pages 1528--1540, 2016.

\bibitem{shen2012object}
Xiaohui Shen, Zhe Lin, Jonathan Brandt, Shai Avidan, and Ying Wu.
\newblock Object retrieval and localization with spatially-constrained
  similarity measure and k-nn re-ranking.
\newblock In {\em 2012 IEEE Conference on Computer Vision and Pattern
  Recognition}, pages 3013--3020. IEEE, 2012.

\bibitem{Song2019Occ}
Lingxue Song, Dihong Gong, Zhifeng Li, Changsong Liu, and Wei Liu.
\newblock Occlusion robust face recognition based on mask learning with
  pairwise differential siamese network.
\newblock In {\em 2019 IEEE/CVF International Conference on Computer Vision
  (ICCV)}, pages 773--782, 2019.

\bibitem{stylianouSimVis2019}
Abby Stylianou, Richard Souvenir, and Robert Pless.
\newblock Visualizing deep similarity networks.
\newblock In {\em IEEE Winter Conference on Applications of Computer Vision
  (WACV)}, 2019.

\bibitem{sun2015deeply}
Yi Sun, Xiaogang Wang, and Xiaoou Tang.
\newblock Deeply learned face representations are sparse, selective, and
  robust.
\newblock In {\em Proceedings of the IEEE conference on computer vision and
  pattern recognition}, pages 2892--2900, 2015.

\bibitem{swearingen2021lookalike}
Thomas Swearingen and Arun Ross.
\newblock Lookalike disambiguation: Improving face identification performance
  at top ranks.
\newblock In {\em 2020 25th International Conference on Pattern Recognition
  (ICPR)}, pages 10508--10515. IEEE, 2021.

\bibitem{szegedy2017inception}
Christian Szegedy, Sergey Ioffe, Vincent Vanhoucke, and Alexander~A Alemi.
\newblock Inception-v4, inception-resnet and the impact of residual connections
  on learning.
\newblock In {\em Thirty-first AAAI conference on artificial intelligence},
  2017.

\bibitem{trigueros2018enhancing}
Daniel~S{\'a}ez Trigueros, Li Meng, and Margaret Hartnett.
\newblock Enhancing convolutional neural networks for face recognition with
  occlusion maps and batch triplet loss.
\newblock {\em Image and Vision Computing}, 79:99--108, 2018.

\bibitem{wan2017occlusion}
Weitao Wan and Jiansheng Chen.
\newblock Occlusion robust face recognition based on mask learning.
\newblock In {\em 2017 IEEE international conference on image processing
  (ICIP)}, pages 3795--3799. IEEE, 2017.

\bibitem{wang2021mlfw}
Chengrui Wang, Han Fang, Yaoyao Zhong, and Weihong Deng.
\newblock Mlfw: A database for face recognition on masked faces.
\newblock {\em arXiv preprint arXiv:2109.05804}, 2021.

\bibitem{wang2012supervised}
Fan Wang and Leonidas~J Guibas.
\newblock Supervised earth mover’s distance learning and its computer vision
  applications.
\newblock In {\em European Conference on Computer Vision}, pages 442--455.
  Springer, 2012.

\bibitem{wang2018cosface}
Hao Wang, Yitong Wang, Zheng Zhou, Xing Ji, Dihong Gong, Jingchao Zhou, Zhifeng
  Li, and Wei Liu.
\newblock Cosface: Large margin cosine loss for deep face recognition.
\newblock In {\em CVPR}, pages 5265--5274, 2018.

\bibitem{wright2008robust}
John Wright, Allen~Y Yang, Arvind Ganesh, S~Shankar Sastry, and Yi Ma.
\newblock Robust face recognition via sparse representation.
\newblock {\em IEEE transactions on pattern analysis and machine intelligence},
  31(2):210--227, 2008.

\bibitem{xu2020improving}
Xiang Xu, Nikolaos Sarafianos, and Ioannis~A Kakadiaris.
\newblock On improving the generalization of face recognition in the presence
  of occlusions.
\newblock In {\em Proceedings of the IEEE/CVF Conference on Computer Vision and
  Pattern Recognition Workshops}, pages 798--799, 2020.

\bibitem{yang2011robust}
Meng Yang, Lei Zhang, Jian Yang, and David Zhang.
\newblock Robust sparse coding for face recognition.
\newblock In {\em CVPR 2011}, pages 625--632. IEEE, 2011.

\bibitem{yi2014learning}
Dong Yi, Zhen Lei, Shengcai Liao, and Stan~Z Li.
\newblock Learning face representation from scratch.
\newblock {\em arXiv preprint arXiv:1411.7923}, 2014.

\bibitem{zhang2020deepemdv2}
Chi Zhang, Yujun Cai, Guosheng Lin, and Chunhua Shen.
\newblock Deepemd: Differentiable earth mover's distance for few-shot learning,
  2020.

\bibitem{Zhang_2020_CVPR}
Chi Zhang, Yujun Cai, Guosheng Lin, and Chunhua Shen.
\newblock Deepemd: Few-shot image classification with differentiable earth
  mover's distance and structured classifiers.
\newblock In {\em IEEE/CVF Conference on Computer Vision and Pattern
  Recognition (CVPR)}, June 2020.

\bibitem{Zhang2016IEEE}
Kaipeng Zhang, Zhanpeng Zhang, Zhifeng Li, and Yu Qiao.
\newblock Joint face detection and alignment using multitask cascaded
  convolutional networks.
\newblock {\em IEEE Signal Processing Letters}, 23(10):1499--1503, 2016.

\bibitem{zhang2020understanding}
Xuanmeng Zhang, Minyue Jiang, Zhedong Zheng, Xiao Tan, Errui Ding, and Yi Yang.
\newblock Understanding image retrieval re-ranking: A graph neural network
  perspective.
\newblock {\em arXiv preprint arXiv:2012.07620}, 2020.

\bibitem{zhao2017robust}
Fang Zhao, Jiashi Feng, Jian Zhao, Wenhan Yang, and Shuicheng Yan.
\newblock Robust lstm-autoencoders for face de-occlusion in the wild.
\newblock {\em IEEE Transactions on Image Processing}, 27(2):778--790, 2017.

\bibitem{zhao2021towards}
Wenliang Zhao, Yongming Rao, Ziyi Wang, Jiwen Lu, and Jie Zhou.
\newblock Towards interpretable deep metric learning with structural matching.
\newblock In {\em ICCV}, 2021.

\bibitem{Tianyue2017calfw}
Tianyue Zheng, Weihong Deng, and Jiani Hu.
\newblock Cross-age {LFW:} {A} database for studying cross-age face recognition
  in unconstrained environments.
\newblock {\em CoRR}, abs/1708.08197, 2017.

\bibitem{zhong2020towards}
Yaoyao Zhong and Weihong Deng.
\newblock Towards transferable adversarial attack against deep face
  recognition.
\newblock {\em IEEE Transactions on Information Forensics and Security},
  16:1452--1466, 2020.

\bibitem{zhou2020AsArcFaceAA}
Shengyao Zhou, Junfan Luo, Junkun Zhou, and Xiang Ji.
\newblock Asarcface: Asymmetric additive angular margin loss for fairface
  recognition.
\newblock In {\em ECCV Workshops}, 2020.

\bibitem{zhou2009face}
Zihan Zhou, Andrew Wagner, Hossein Mobahi, John Wright, and Yi Ma.
\newblock Face recognition with contiguous occlusion using markov random
  fields.
\newblock In {\em 2009 IEEE 12th international conference on computer vision},
  pages 1050--1057. IEEE, 2009.

\end{thebibliography}
